\documentclass[10pt,twocolumn,letterpaper]{article}

\usepackage{iccv}
\usepackage{times}
\usepackage{epsfig}
\usepackage{graphicx}
\usepackage{amsmath}
\usepackage{amssymb}

\usepackage{booktabs}
\usepackage{appendix}
\usepackage{amssymb}
\usepackage[vlined,ruled,linesnumbered]{algorithm2e}
\usepackage{multirow}

\usepackage[breaklinks=true,bookmarks=false, pagebackref=true]{hyperref}
\hypersetup{
	colorlinks=true,
	linkcolor=red,
        anchorcolor=red,
	filecolor=blue,      
	urlcolor=red,
	citecolor=green,
}

\iccvfinalcopy % *** Uncomment this line for the final submission

\def\iccvPaperID{3908} % *** Enter the ICCV Paper ID here

\ificcvfinal\fi

\begin{document}

%%%%%%%%% TITLE
\title{LiDAR-Camera Panoptic Segmentation via Geometry-Consistent and Semantic-Aware Alignment}

\author{
Zhiwei Zhang$^{1\dagger}$
\quad Zhizhong Zhang$^{2,3\dagger}$
\quad Qian Yu$^2$
\quad Ran Yi$^1$
\quad Yuan Xie$^{2,3*}$
\quad Lizhuang Ma$^{1,2*}$\\
$^1$School of Electronic Information and Electrical Engineering, Shanghai Jiaotong Univeristy, China\\
$^2$Department of Computer Science and Engineering, East China Normal University, China\\
$^3$Chongqing Institute of East China Normal University, China\\
{\tt\small \{zhangzw12319,  ranyi,  lzma\}@sjtu.edu.cn, \{qianyu, zzzhang, yxie\}@cs.ecnu.edu.cn}
% \and
}

\maketitle
{
\renewcommand{\thefootnote}%
    {\fnsymbol{footnote}}
  \footnotetext[1]{Corresponding authors; $\dagger$ equal contributions}
}
\ificcvfinal\thispagestyle{empty}\fi

%%%%%%%%% ABSTRACT
\begin{abstract}
    3D panoptic segmentation is a challenging perception task that requires both semantic segmentation and instance segmentation. In this task, we notice that images could provide rich texture, color, and discriminative information, which can complement LiDAR data for evident performance improvement, but their fusion remains a challenging problem. To this end, we propose LCPS, the first LiDAR-Camera Panoptic Segmentation network. In our approach, we conduct LiDAR-Camera fusion in three stages: 1) an Asynchronous Compensation Pixel Alignment (ACPA) module that calibrates the coordinate misalignment caused by asynchronous problems between sensors; 2) a Semantic-Aware Region Alignment (SARA) module that extends the one-to-one point-pixel mapping to one-to-many semantic relations; 3) a Point-to-Voxel feature Propagation (PVP) module that integrates both geometric and semantic fusion information for the entire point cloud. Our fusion strategy improves about $6.9\%$ PQ performance over the LiDAR-only baseline on NuScenes dataset. Extensive quantitative and qualitative experiments further demonstrate the effectiveness of our novel framework. The code will be released at \href{https://github.com/zhangzw12319/lcps.git}{https://github.com/zhangzw12319/lcps.git}.
\end{abstract}

%%%%%%%%% BODY TEXT
\section{Introduction}\label{sec:intro}

3D scene perception has become an increasingly important task for a wide range of applications, including self-driving and robotic navigation. Lying in the heart of 3D vision, 3D panoptic segmentation is a comprehensive perception task composed of semantic and instance segmentation \cite{kirillov2019panoptic}. This is still challenging since it not only requires predicting semantic labels of each point for \textit{Stuff} classes, such as \textit{tree}, \textit{road}, but also needs recognizing instances for \textit{Thing} classes, e.g., \textit{car}, \textit{bicycle}, and \textit{pedestrian} simultaneously.

 \begin{figure}[!htbp]
 \centering
 \includegraphics[width=0.45\textwidth]{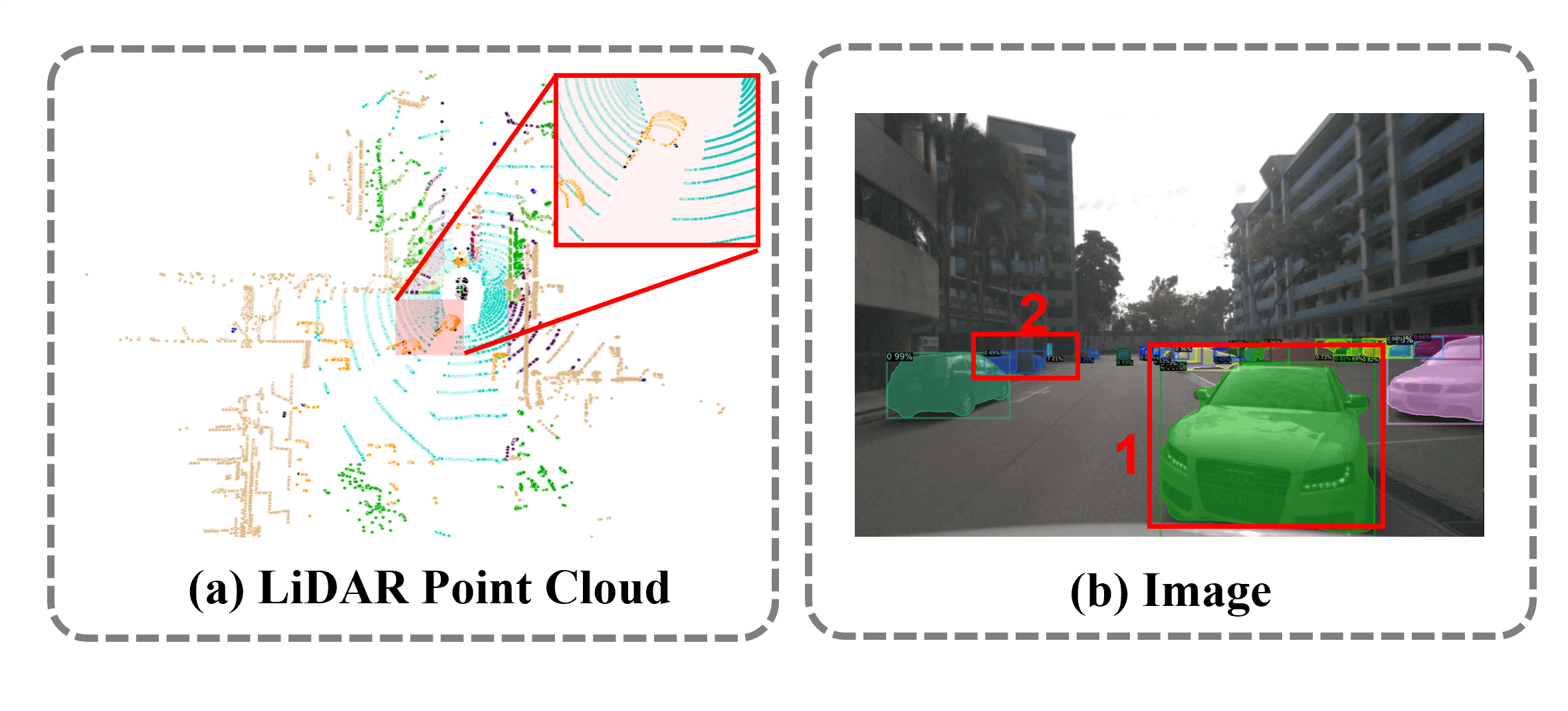} % Reduce the figure size so that it is slightly narrower than the column.
 \caption{
 The distinctions between LiDAR point cloud and images. (a) The red box displays a vehicle segment (orange points) in the point cloud, where points are sparsely and unevenly distributed. (b) The lower-right green mask demonstrates a vehicle with dense texture and color features, effectively detected via \cite{zhou2021probablistic}. The upper-left blue mask (partly occluded) shows image features that help detect small objects in the distance. Better zoomed in.}
 \label{fig:sec1_difference}
 \end{figure}

Currently, the leading 3D panoptic methods use LiDAR-only data as input sources. However, We have observed that using only LiDAR data for perception has some insufficiencies: 1) LiDAR point cloud is usually sparse and unevenly distributed, as illustrated in Figure \ref{fig:sec1_difference} (a), making it challenging for 3D networks to capture the notable difference between the foreground and the background; 2) distant objects that occupy just a few points appear to be small in the view and cannot be effectively detected. On the contrary, images provide rich texture and color information, as shown in Figure \ref{fig:sec1_difference} (b). This observation motivates us to use images as an additional input source to complement LiDAR sensors for scene perception. Moreover, most autonomous driving systems come equipped with RGB cameras, which makes LiDAR-Camera fusion studies more feasible.

Although LiDAR sensors and cameras complement each other, their fusion strategy remains challenging. Existing fusion strategy could be generally split into proposal-level fusion \cite{Ku_2018}, result-level fusion \cite{Qi_2018}, and point-level fusion \cite{Vora_2020, Huang_2020, Wang_2021}, as summarized in PointAugmenting \cite{wang2021pointaugmenting}. Yet, proposal-level and result-level fusion focus on integrating 2D and 3D proposals (or bounding box results) for object detection, which limits their generalizability in dense predictions like segmentation tasks.
   The previous point-fusion methods also suffer: 
   1) the asynchronous working frequency between LiDAR and camera sensors is not considered, which may result in misaligned feature correspondence;
   2) point-fusion is a one-to-one fusion mechanism, and large image areas are unable to be mapped to sparse LiDAR points, resulting in the waste of abundant information from dense pixel features; \textit{e.g.}, for a 32-beams LiDAR sensor, only about $5\%$ pixels can be mapped to correlated points, while the $95\%$ of pixel features would be dropped \cite{liu2023bevfusion}.
   3) previous point-level fusion methods \cite{Vora_2020, Huang_2020, Wang_2021} often use simple concatenation, which excludes points whose projections fall outside the image plane, as image features cannot support them.

Motivated by these insufficiencies, we propose the first LiDAR-Camera Panoptic Segmentation (LCPS) network to exploit the complementary information from multiple sensors. In this work, we propose a novel three-stage fusion strategy involving the Asynchronous Compensation Pixel Alignment (ACPA) module, Semantic-Aware Region Alignment (SARA) module, and Point-to-Voxel feature Propagation (PVP) module. 
The ACPA module employs ego-motion compensation operations to achieve spatial-temporal alignment between the LiDAR and camera modalities, overcoming asynchronous issues in point fusion. 
Then, our novel SARA module extends the one-to-one point-pixel mapping to one-to-many semantic relations, highly improving the image utilization rate. Specifically, SARA introduces Class Activation Maps (CAM) for image branch to localize semantic-related image regions for each point.
Next, the PVP module replaces simple concatenation with local attention to propagate information from point-aligned pixels and regions to the entire point cloud. Points outside camera frustums can also be preserved and attached to image features. Finally, we design a Foreground Object selection Gate (FOG) module to enforce the network to learn a class-agnostic foreground object mask in addition to the semantic prediction head. This gate effectively reduces incorrect predictions and stabilizes the training process. To sum up, our main contributions are:

\begin{itemize}
    \item To the best of our knowledge, this is the first LiDAR-Camera fusion network for 3D panoptic segmentation, which effectively exploits the complementary information of the LiDAR and image data.
    
    \item We have improved the former point-fusion approach with our novel Asynchronous Compensation Pixel Alignment (ACPA), Semantic-Aware Region Alignment (SARA), and Point-to-Voxel feature Propagation (PVP) modules. These contribute to the geometry-consistent and semantic-aware alignment between LiDAR and Camera sensors.
    
    \item We present the Foreground Object selection Gate (FOG) to reduce the incorrect predictions of confusing points, further boosting panoptic segmentation quality.
    
    \item Extensive quantitative and qualitative experiments demonstrate the effectiveness of our approach. Our fusion approach improves performance at $6.9\%$ PQ on NuScenes and $3.3 \%$ PQ in SemanticKITTI compared to the LiDAR-only baseline.
\end{itemize}

\begin{figure*}[!htbp]
\centering
\includegraphics[width=1.0\textwidth]{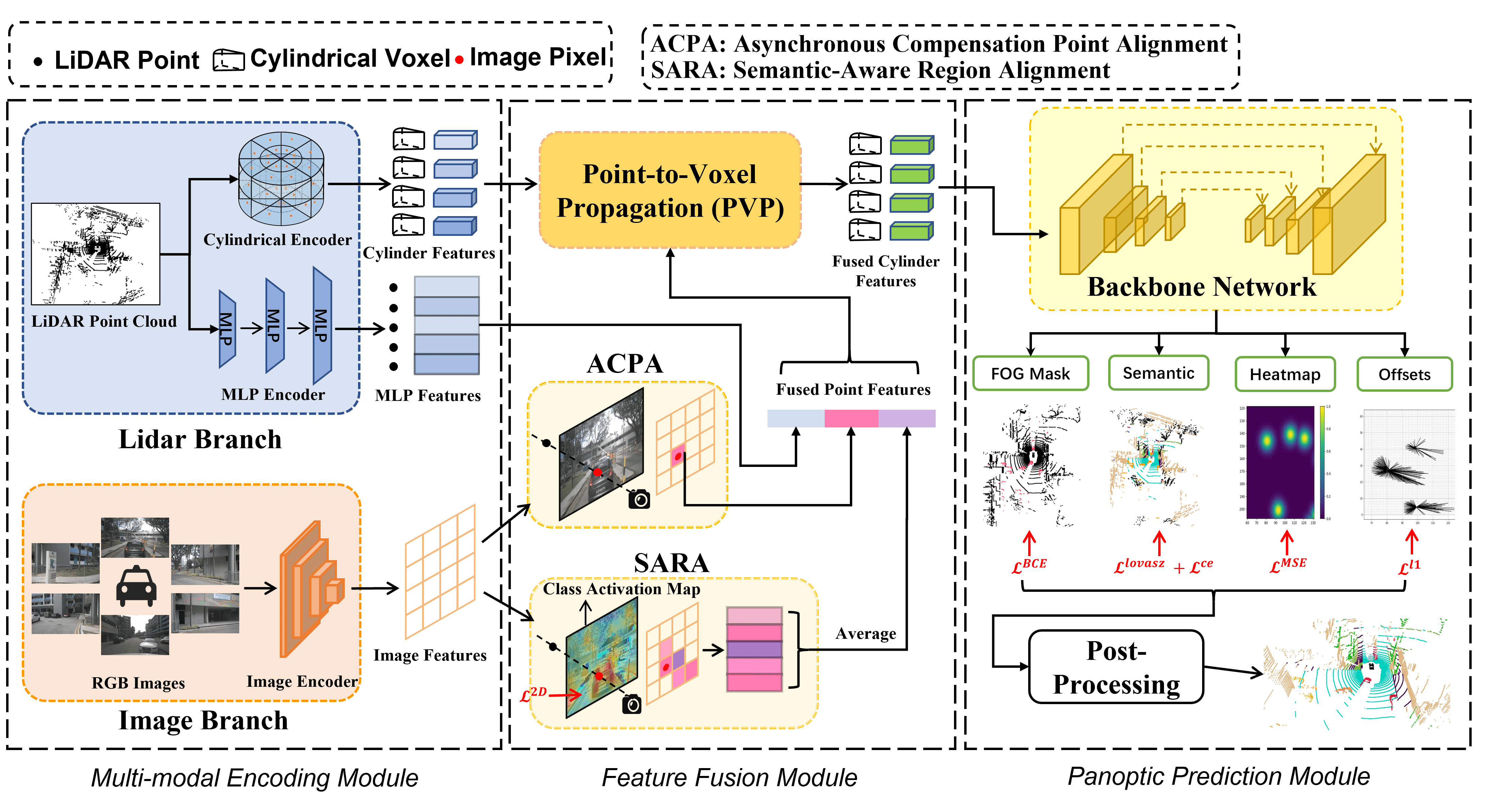} % Reduce the figure size so that it is slightly narrower than the column.
\caption{The overall pipeline of our LiDAR-Camera Panoptic Segmentation network (LCPS). LCPS consists of multi-modal encoding, feature fusion, and panoptic prediction modules. The encoding module extracts cylinder features, MLP features, and image features. In the fusion stage, MLP features are geometrically and semantically aligned with pixel features via ACPA and SARA. Next, the PVP module merges fused point features with original cylinder features to obtain fused ones. Finally, the panoptic prediction module yields predictions of four heads, which are post-processed to obtain panoptic segmentation results.}
\label{fig2:overview}
\end{figure*}

%-------------------------------------------------------------------------
\section{Related Work} \label{sec:realtaed_work}

Panoptic segmentation is initially proposed from 2D vision \cite{kirillov2019panoptic}, for the purpose of integrating semantic and instance segmentation. Later, research of panoptic segmentation extends to videos and LiDAR point cloud. Early work LPSAD \cite{milioto2020lidar} handles LiDAR panoptic segmentation via projecting points into range view and then using 2D convolution network to extract features. Although pure 2D network can boost efficiency, it also suffers performance degradation when mapping 2D predictions back to the point cloud. Later, 3D LiDAR networks are designed for this task. Generally, 3D panoptic segmentation can be divided into two categories, {\it i.e.,}  proposal-based and proposal-free methods.

\noindent \textbf{Proposal-based 3D Panoptic Segmentation.} Proposal-based methods Panoptic-Deeplab \cite{Cheng_2020} and EfficientLPS \cite{sirohi2021efficientlps} predict bounding-box proposals and then merge them with semantic results to obtain panoptic predictions, following classical object detection framework\cite{chen2017deeplab, he2017mask}. 
However, proposal-based methods tend to result in inconsistent segmentation between instance and semantic branches. Moreover, the segmentation result is susceptible to the quality of object detection.

\noindent \textbf{Proposal-free 3D Panoptic Segmentation.} 
Proposal-free methods abandon object proposals and predict object center and point offset instead. The post-processing module will cluster points into instance groups according to object center and point offset.
DS-Net \cite{DBLP:conf/cvpr/Hong0Z0L21} proposes a dynamic-shifting mechanism of instance points toward its possible centers for Mean Shift clustering. SMAC-Seg \cite{Li_2022} and SCAN \cite{xu_2022} attempt to use attention module on multi-directional or multi-scale feature maps. GP-S3Net \cite{razani_gp-s3net_2021} constructs a dynamic graph composed of foreground clusters as graph nodes, processed by graph convolutional network for instance segmentation branch. Panoptic-Polarnet \cite{zhou_panoptic_polarnet_2021} projects 3D features into BEV and utilizes learnable BEV heatmap with non-maximum suppression(NMS) to predict centers. Following Panoptic-Polarnet's BEV design, Panoptic-PHNet \cite{li2022panoptic} improves center and offset generation by replacing NMS with a center grouping module to merge duplicated centers, as well as augmenting offset via KNN-Transformer. For now, Panoptic-PHNet has achieved 1st place on NuScenes and SOTA performance on SemanticKITTI benchmarks.

Nevertheless, sparse and uneven LiDAR points will impose large variance for center and offset predictions in Bird-Eye-View and thus becomes a bottleneck for current SOTA approaches. RGB images can compensate for LiDAR features, which motivates us to design LCPS.

\noindent \textbf{LiDAR-Camera Fusion Models.}
In object detection and semantic segmentation, pioneering research already considers modal fusion between images and LiDAR points. For example, PMF \cite{zhuang2021perception} attempts to project LiDAR points to the perspective view and proposes a two-branch 2D network to extract semantic features with an attentive fusion module. TransFuser \cite{prakash2021multi} and TransFusion \cite{bai2022transfusion} 
consider utilizing transformers to fuse 3D LiDAR points and 2D images. DeepFusion \cite{li2022deepfusion} focuses on how to avoid feature misalignment when extensive data augmentation is performed in both LiDAR and camera branches. However, multi-modal panoptic segmentation has yet to be explored, accompanied by asynchronous and utilization issues. 

%-------------------------------------------------------------------------
\section{Methodology} \label{sec:methodology}

\subsection{Overview} \label{sec:overview}

\noindent \textbf{Problem Formulation.}
This paper considers 3D panoptic segmentation \cite{fong2022panoptic}. Formally, we denote a set of LiDAR points as $\{(x_i^{\text{3D}}, f_i^{\text{3D}}) | (x_i^{\text{3D}} \in \mathbb{R}^3, f_i^{\text{3D}} \in \mathbb{R}^C \}_{i=1}^N$, where $N$, $x_i^{\text{3D}}$ and $f_i^{\text{3D}}$ represent the total number of points, 3D positions, and point features of $C$ dimensions, respectively. This task requires predicting a unique semantic class $\{\hat{y}_i^{\text{3D}}\}_{i=1}^N$ for each point and accurately identifying groups of points as foreground objects with an instance ID, denoted as $\{\text{ID}^{\text{3D}}_i\}_{i=1}^N$. 

Besides, we assume that $K$ surrounding cameras, which are cheap and common, capture images associated with the LiDAR frame for LiDAR-Camera fusion. Similarly, we represent each image as a set of pixels $\{ (x_{k,i}^{\text{2D}}, f_{k,i}^{\text{2D}}) | (x_{k,i}^{\text{2D}} \in \mathbb{R}^2, f_{k,i}^{\text{2D}} \in \mathbb{R}^C \}_{i=1, k=1}^{i=N', k=K}$, where $N'$, $x_i^{\text{2D}}$, $f_i^{\text{2D}}$ and $k$ represent the total number of pixels, 2D positions, pixel features, and the camera index, respectively. Our primary objective in this paper is to improve panoptic segmentation performance by fully exploring the complementary information in the LiDAR and Camera sensors.

\noindent \textbf{Pipeline Architecture.}
The framework in Figure \ref{fig2:overview} consists of a multi-modal encoding module, a LiDAR-Camera feature fusion module, and a panoptic prediction module. In the encoding stage, the LiDAR points are respectively encoded by a cylindrical voxel encoder and an MLP encoder, while the images are encoded using SwiftNet \cite{Wang_2021_CVPR}.
%The MLP and image features, which are not strictly correlated, {\it i.e.,} points and pixels are not one-to-one mapped, are then aligned by the proposed asynchronous compensation point alignment and semantic-aware region alignment.
In the fusion stage, the MLP feature and image features, which are not strictly correlated, are first aligned through the proposed Asynchronous Compensation and Semantic-Aware Region Alignment, and then are concatenated into fused point features. Subsequently, our Point-to-Voxel Propagation module (PVP) accepts the fused point features and outputs the final cylinder representation.
In the prediction stage, the backbone network includes a proposed FOG head, a semantic segmentation head, a heatmap head, and an offsets head. The latter two heads follow Panoptic-Polarnet \cite{zhou_panoptic_polarnet_2021}, where we regress a binary object center mask and a 2D offset among bird-eye-view grids. During inference, the post-processing shifts the predicted foreground BEV grids to their nearest centers and clusters the points within the grids into instances.

\subsection{Asynchronous Compensation Pixel Alignment} \label{sec:ACPA}

A straightforward solution \cite{Liang_2018,Vora_2020, Zhuang_2021} for fusing LiDAR and Camera is to establish point-to-pixel mappings, such that points can be directly projected to image planes and decorated with pixel features. However, this mapping would lead to false mapping due to the asynchronous frequency between cameras and LiDAR sensors. For instance, on NuScenes dataset, each camera operates at a frequency of 12Hz, while the LiDAR sensor operates at 20Hz.

Motivated by this, we improve the point-level fusion by incorporating additional asynchronous compensation to achieve a consistent geometric alignment over time. The fundamental idea is to transform the LiDAR points into a new 3D coordination system when the corresponding images are captured at that time. The transformation matrix is obtained by considering the ego vehicle's motion matrix. Specifically, let $t_1$ and $t_2$ denote the time when the LiDAR point cloud and the related images are captured. Then we have:

\textbf{Step-1.} Transform LiDAR points from world coordinates to ego-vehicle coordinates at time $t_1$. By multiplying the coordinate transformation matrix $\mathbf{T}_{t_1}^{\text{W} \rightarrow \text{V}} $ provided by the dataset, we can obtain the 3D position in the ego-vehicle coordinate system, denoted as  $\hat{x}_i^{\text{3D}}$.

\textbf{Step-2.} Transform LiDAR points in ego-vehicle coordinates from time $t_1$ to time $t_2$. %for ego-motion compensation. 
To achieve this, a time-variant transformation matrix is required, denoted $\mathbf{T}_{t_1 \rightarrow t_2}^{\text{V} \rightarrow \text{V}}$. However, such a matrix is often not directly available in datasets. Instead, the ego vehicle's motion matrices from the current frame to the first frame are often provided for each sliced sequence. Therefore, we can divide $\mathbf{T}_{t_1 \rightarrow t_2}^{\text{V} \rightarrow \text{V}}$ as the product of $ (\mathbf{T}_{t_2 \rightarrow t_0}^{\text{V} \rightarrow \text{V}})^{-1}$ and $ \mathbf{T}_{t_1 \rightarrow t_0}^{\text{V} \rightarrow \text{V}} $, where $t_0$ is the time of the first frame. Using this ego-motion transformation matrix, we obtain the point position in ego-vehicle coordinates at time $t_2$, denoted as  $\tilde{x}_i^{\text{3D}}$.

\textbf{Step-3.} Obtain pixel features at time $t_2$. By using camera extrinsic and intrinsic matrices ($\mathbf{E}_k$ and $\mathbf{I}_k$), we get the projected 2D position $\tilde{x}_{k, i}^{\text{2D}}$ of each point in the $k_{\text{th}}$ image plane at time $t_2$. After excluding the points whose projections are outside the image plane, the resulting pixel features $\{\tilde{f}_{k,i}^{\text{2D}}\}_{i=1}^{N_k}$ are indexed by $\tilde{x}_{k,i}^{\text{2D}}$. $N_k$ is the number of points inside the image plane (${N_k} < N$).

These homogeneous transformation steps can be summarized in the following equation:

\begin{equation}
\label{proj_eq}
\left[ \begin{array}{c} \tilde{x}_{k,i}^{\text{2D}} \\ 1 \end{array} \right] = \mathbf{I}_k \mathbf{E}_k\mathbf{T}_{t_1 \rightarrow t_2}^{\text{V} \rightarrow \text{V}} \mathbf{T}_{t_1}^{\text{W} \rightarrow \text{V}}\left[\begin{array}{c} x_i^{\text{3D}} \\ 1 \end{array} \right].
\end{equation}

In summary, we obtain pixel-aligned features for each point using Equation \ref{proj_eq}. Our approach adopts ego-motion compensation via Step 2, resulting in a simple but more accurate geometric-consistent feature alignment.

\subsection{Semantic-Aware Region Alignment} \label{sec:SARA}

\begin{figure}[t]
 \centering
 \includegraphics[width=0.5\textwidth]{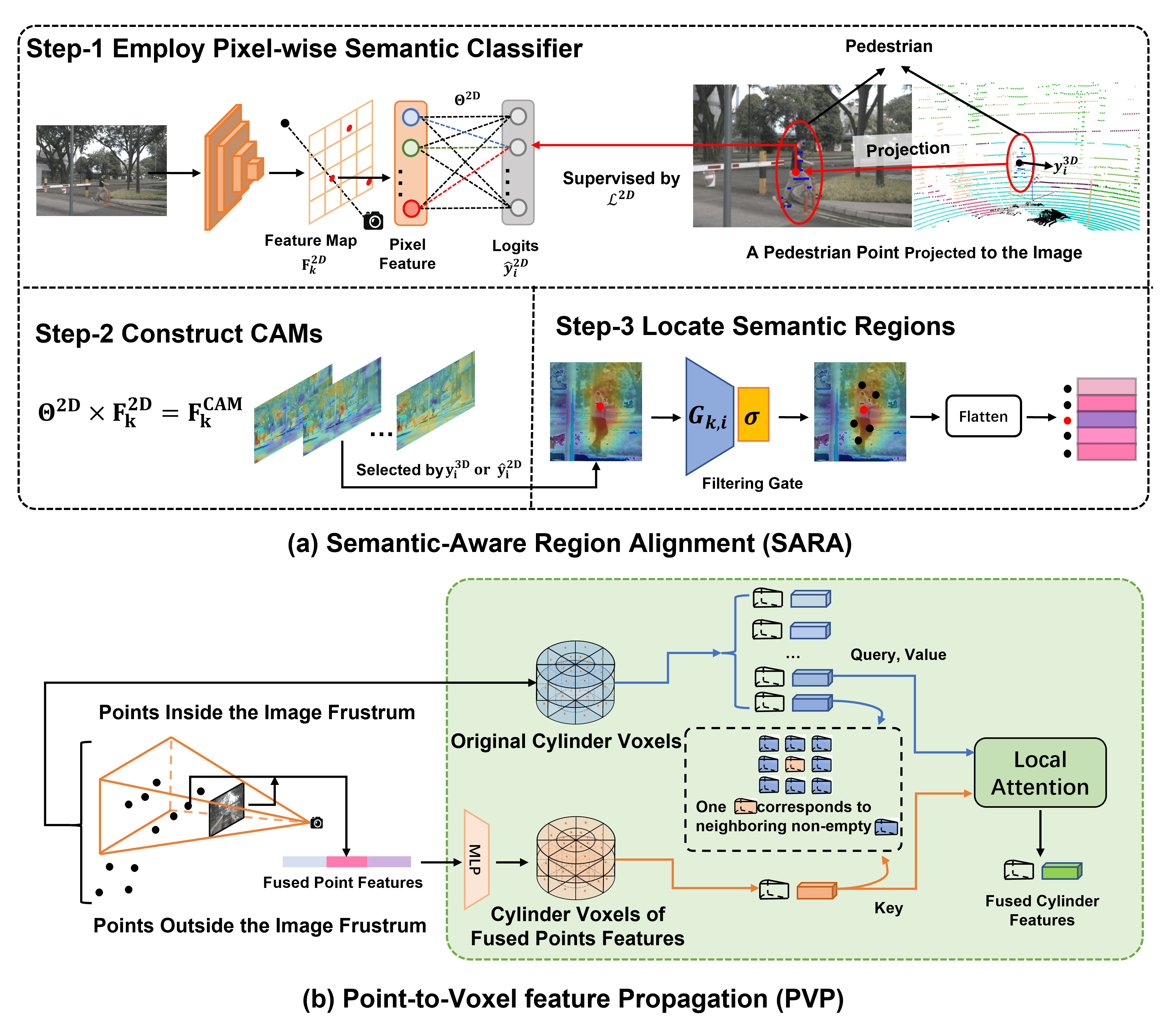} % Reduce the figure size so that it is slightly narrower than the column.
 \caption{(a) Overview of the SARA module, which employs pixel-wise semantic classifier, constructs CAMs and locates semantic regions. 
 (b) Overview of the PVP Module, which involves a cylindrical partition of fused point features and attentive propagation. Better zoomed in. }
 \label{fig:sec34}
\end{figure}

 Due to the sparse nature and limited eyeshot of LiDAR point clouds, only a small fraction of image features can be matched with LiDAR points. To address this issue, we propose to find semantic-relevant regions, extending the one-to-one mapping to one-to-many relations. Inspired by \textit{Class Activation Maps} (CAMs) \cite{Zhou_2016, mcever2020pcams}, we present a Semantic-Aware Region Alignment module by using image CAMs to localize relevant semantic regions, as illustrated in Figure \ref{fig:sec34} (a).

\textbf{Step-1.} %Using point labels to learn an image classifier. 
We first introduce a pixel-wise semantic classifier $\phi^{\text{2D}}(\cdot)$ to learn the semantic information in the image branch, and define $\mathbf{\Theta^{\text{2D}}}\in \mathbb{R}^{M \times C}$ as the classifier parameters, where $M$ is the number of semantic categories. Based on the observation that projected pixels share the same semantic category with matched points, we use point labels to train the image classifier with cross-entropy loss: 

\begin{equation}
    \label{eq:L_2d}
    \mathcal{L}_{\text{2D}} = -\frac{1}{N_k}\sum_{i=1}^{N_k} y_{i}^{\text{3D}} \log(\hat{y}_{i}^{\text{2D}}),
\end{equation}

\noindent where $\hat{y}_{i}^{\text{2D}}$ and $y_{i}^{\text{3D}}$ denote the predicted pixel label and related ground-truth point label (such alignment is obtained in Section \ref{sec:ACPA}), and $N_K$ represents the number of points which can be projected into the $k$-th image plane.   

\textbf{Step-2.} We use this classifier to generate the class activation maps (CAMs). Let $ \mathbf{F}_k^{\text{2D}} \in \mathbb{R}^{ C \times H^{\text{2D}} \times W^{\text{2D}} }$ be the image feature map extracted by the last convolution layer, and $H^{\text{2D}}$ and $W^{\text{2D}}$ are the height and width of image feature maps. We can then obtain CAMs using the following formula:

\begin{equation}
    \label{eq:F_CAM}
    \mathbf{F}_k^{\text{CAM}} = \mathbf{\Theta}^{\text{2D}} \times  \mathbf{F}_k^{\text{2D}},
\end{equation}

\noindent where $\times$ denotes the matrix multiplication. The generated CAMs are represented by $\mathbf{F}_k^{\text{CAM}} \in \mathbb{R}^{ M \times H^{\text{2D}} \times W^{\text{2D}}}$. Each channel in CAM is a $ H^{\text{2D}} \times W^{\text{2D}}$ heatmap related to a specific semantic category.

\textbf{Step-3.} For each LiDAR point, we use the generated CAMs to localize sets of pixels as semantic-related image regions. We design a filtering gate $\mathbf{G}_{k,i}^y \in \mathbb{R}^{H^{\text{2D}} \times W^{\text{2D}}}$, constructed by selecting a single heatmap of class $y$ from CAMs $F_k^{\text{CAM}}$ according to the ground-truth or predicted pixel label. The gate is controlled by subtracting a predefined confidence threshold $\tau$. 
Pixels with heatmap values lower than that threshold will be set to zero in $\mathbf{G}_{k,i}^y$. Finally, we get a set of related pixels:

\begin{equation}
    \label{eq:pixel_region}
    \{ f^{\text{CAM}} \}_{k,i} =  \textit{Flatten}(\sigma(\mathbf{G}_{k,i}^y \otimes \mathbf{F}_k^{\text{2D}})),
\end{equation}

\noindent where $\otimes$ denotes element-wise multiplication, and $\sigma$ denotes the activation function. 
\textit{Flatten} function is adopted to transform features from matrix format $C \times H^{\text{2D}} \times W^{\text{2D}}$ into a set format $(H^{\text{2D}} W^{\text{2D}}) \times C$, followed by discarding zero vectors which is filtered by $G_{k,i}$.
Consequently, we obtain a set of pixel features $ \{ f^{\text{CAM}} \in \mathbb{R}^{C} \}_{k,i} $ for each LiDAR point $i$ and each camera $k$.

We finally average the set of region features to a single vector, then concatenate it with the MLP output and pixel-aligned features to constitute the fused point features.
In summary, unlike one-to-one pixel alignment via pure geometric projection, the image regions are directly collected in a one-to-many semantic-aware manner.

\subsection{Point-to-Voxel Feature Propagation} \label{sec:PVP}
Image features seem to not support the points outside the camera frustum; therefore, these points are usually excluded \cite{Sindagi_2019, Vora_2020, Huang_2020}. To overcome this problem, we propose the Point-to-Voxel Feature Propagation to integrate both geometric and semantic information for the entire point cloud. To this end, we choose cylindrical voxels as the bridge to complete the fusion process since the tensor shape of the voxel representation is invariant to the alteration of point numbers, which naturally provides an alignment between the original point cloud and the image-related point cloud subset.

 As shown in Figure \ref{fig:sec34} (b), a cylindrical encoder first encodes the original point cloud into voxels. %while point features and the other two types of pixel features are concatenated. 
 Meanwhile, for the fused point features, we first align their channel dimensions with the original voxel using MLP, and then divide these fused points into another set of cylindrical voxels, where features will be scattered and pooled within the same voxel to obtain voxel features. A noticeable observation is that a LiDAR point may have alignment with more than one camera, resulting in multiple fused point features of a single point. Therefore, we treat such multiple features as multiple points at the same 3D position during voxelization. Then we propagate the voxels of the fused point features (denoted as $\vartheta^{\text{im}}$) to the original cylindrical voxels (denoted as $\vartheta^{\text{p}}$) using modified local attention \cite{vaswani2017attention}. In this attention mechanism, each voxel $\vartheta^{\text{p}}$ acts as queries $Q$, while the neighboring $27$ $\vartheta^{\text{im}}$ voxels act as keys $K$ and values $V$. Then the computation is given by:
 
 \begin{equation}
     \text{Att}(\vartheta^{\text{p}}, \vartheta^{\text{im}}, \vartheta^{\text{im}}) = \text{Softmax}(\frac{QK^T}{\sqrt{C}})V,
 \end{equation}
 
\noindent where $C$ is the channel dimensions. After that, we add the attentive voxels with original $\vartheta^{\text{p}}$ to make a residual connection, as shown in the following equation:

\begin{equation}
    \label{eq:add}
    \vartheta = \text{Att}(\vartheta^{\text{p}}, \vartheta^{\text{im}}, \vartheta^{\text{im}}) + \vartheta^{\text{p}}.
\end{equation}
 
 Through this attentive propagation, information from the entire point cloud and multiple cameras is comprehensively integrated into a single cylindrical voxel representation $\vartheta$.

\subsection{Improved Panoptic Segmentation} \label{sec:improved_ps}

Here we briefly describe the Foreground Object selection Gate (FOG) head and loss functions for panoptic prediction. Other implementation details are displayed in Section \ref{sec:implementation} and the Appendix.

\noindent \textbf{Foreground Object Selection Gate.}
  In Panoptic-PolarNet \cite{zhou_panoptic_polarnet_2021}, the panoptic network diverges into three prediction heads for semantic labels, centers, and offset prediction. However, we find that semantic predictions largely affect the final quality of panoptic segmentation. This is because the center and offset head only provide class-agnostic predictions, while accurate semantic information is required for post-processing to cluster foreground grids into the nearest object centers. Inspired by \cite{lim2020learning}, we propose FOG, a Foreground Object Selection Gate, to enhance the original semantic classifier. FOG is a binary classifier aiming to differentiate foreground objects. Given voxel features obtained from the backbone network as $\vartheta^{\text{b}} \in \mathbb{R}^{ H\times W \times Z}$, FOG predicts a class-agnostic binary mask $y^{\text{FOG}} \in [0, 1]^{H\times W \times Z}$, which is supervised by binary cross-entropy loss $\mathcal{L}^{\text{BCE}}$. As a result, the foreground mask complements the semantic head by filtering out background points in the post-processing period.

\noindent \textbf{Loss Designs.}
    The total loss is derived in the following equation: %\textbf{\ref{eq:total_loss}}
    \begin{equation}
        \label{eq:total_loss}
        \mathcal{L}^{\text{total}} = 
        \alpha_1 (\mathcal{L}^{\text{CE}} + \mathcal{L}^{\text{Lovasz}}) + 
        \alpha_2 \mathcal{L}^{\text{MSE}} + 
        \alpha_3 \mathcal{L}^{\text{L1}} + 
        \alpha_4 \mathcal{L}^{\text{BCE}} + 
        \alpha_5 \mathcal{L}^{\text{2D}}.
    \end{equation}

    The top four terms are based on Panoptic-Polarnet \cite{zhou_panoptic_polarnet_2021}. $\mathcal{L}^{\text{CE}}$ and $\mathcal{L}^{\text{Lovasz}}$ represent cross-entropy loss and Lovasz loss \cite{berman2018lovasz} for semantic supervision. $\mathcal{L}^{\text{MSE}}$ is a Mean-Squared-Error (MSE) loss for BEV center heatmap regression. $\mathcal{L}^{\text{L1}}$ is an L1 loss for BEV offset regression. %Further details of these terms will be provided in the Appendix. 
    In addition, the last two terms are new in this paper. $\mathcal{L}^{\text{BCE}}$ represents a binary entropy loss used for FOG head, and $\mathcal{L}^{\text{2D}}$ is a pointly-supervised loss for region-fusion, given by Equation \ref{eq:L_2d}. $\alpha_2$ and $\alpha_3$ are set to 100 and 10 respectively, while the other three weights are set to 1.

\section{Experiments} \label{sec:experiments}
\begin{table*}[htbp]
\scriptsize
\centering
\resizebox{1.0\textwidth}{!}{
% \scalebox{0.65}{
\begin{tabular}{l|cccc|ccc|ccc|c} %需要12列
\toprule %添加表格头部粗线
Method & PQ & PQ$^\dagger$ & SQ & RQ & PQ$^{\textit{th}}$ & SQ$^{\textit{th}}$ & RQ$^{\textit{th}}$ & PQ$^{\textit{st}}$ & SQ$^{\textit{st}}$ & RQ$^{\textit{st}}$ & mIoU \\
%  %有n个&，就表示该行有n+1列
\hline %绘制一条水平横线
 DS-Net \cite{DBLP:conf/cvpr/Hong0Z0L21} & 42.5 & 51.0 & 83.6 & 50.3& 32.5 & 83.1 & 38.3 & 59.2 & 84.4 & 70.3 & 70.7 \\ % √
 GP-S3Net \cite{razani_gp-s3net_2021} & 48.7 & 60.3 & 61.3 & 63.7 &  61.6& 86.4 & 71.7 & 43.8 & 51.8 & 60.8 & 61.8 \\ % √
 PanopticTrackNet \cite{hurtado2020mopt}& 50.0 & 57.3 & 80.9 & 60.6& 45.1 & 80.3& 52.4& 58.3 &81.9 & 74.3& 63.1 \\ % √
 EfficientLPS \cite{sirohi2021efficientlps}& 62.0 & 65.6& 83.4& 73.9& 56.8& 83.2& 68.0& 70.6& 83.8& 83.6 & 65.6 \\ % √
 SCAN \cite{xu_2022}& 65.1 & 68.9 & 85.7 & 75.3& 60.6& 85.7& 70.2& 72.5& 85.7& 83.8& 77.4\\ % √
 Panoptic-PolarNet \cite{zhou_panoptic_polarnet_2021}& 67.7 & 71.0 & 86.0 & 78.1 & 65.2 & 87.2& 74.0 & 71.9 & 83.9& 84.9 & 69.3 \\ % √
 SMAC-Seg HiRes \cite{Li_2022}& 68.4 & 73.4 & 85.2 & 79.7 & 68.0 & 87.3 & 77.2 & 68.8 & 83.0 & 82.1 & 71.2 \\ % √
 CPSeg HR \cite{Li_2023}& 71.1 & 75.6& 85.5& 82.5& 71.5& 87.3& 81.3& 70.6& 83.6& 83.7& 73.2\\ % √
 Panoptic PH-Net \cite{li2022panoptic}& 74.7 & 77.7 & 88.2 & 84.2& 74.0 & 89.0 & 82.5 & \textbf{75.9} & \textbf{86.8} & \textbf{86.9} & 79.7\\ % √
 PUPS \cite{Su_2023} & 74.7 & 77.3 & 89.4 & 83.3 & 75.4 & 91.8 & 81.9 & 73.6 & 85.3 & 85.6 & -\\
 \hline
 \textbf{LCPS (Baseline)} & 72.9 & 77.6 & 88.4 & 82.0 & 72.8 & 90.1 & 80.5 & 73.0 & 85.5 & 84.5 & 75.1 \\
 \textbf{LCPS (Full)} & \textbf{79.8} & \textbf{84.0} & \textbf{89.8} & \textbf{88.5} & \textbf{82.3} & \textbf{91.7} & \textbf{89.6} & 75.6 & 86.7 & 86.5 & \textbf{80.5}\\
          
\bottomrule %添加表格底部粗线
\end{tabular}
%} %end scalebox
}
\setlength{\abovecaptionskip}{2pt}
\caption{3D panoptic segmentation results on NuScenes validation set. The evaluation metric is provided in PQ\%.}
\label{nusc_val}
\end{table*}

\begin{table*}[!ht]
\scriptsize
\centering
\resizebox{1.0\textwidth}{!}{
\begin{tabular}{l|cccc|ccc|ccc|c} %需要11列
\toprule %添加表格头部粗线
Method & PQ & PQ$^\dagger$ & SQ & RQ & PQ$^{\textit{th}}$ & SQ$^{\textit{th}}$ & RQ$^{\textit{th}}$ & PQ$^{\textit{st}}$ & SQ$^{\textit{st}}$ & RQ$^{\textit{st}}$ & mIoU\\
 %有n个&，就表示该行有n+1列
\hline %绘制一条水平横线
 EfficientLPS \cite{sirohi2021efficientlps}& 62.4 & 66.0& 83.7 & 74.1& 57.2& 83.6& 68.2 & 71.1& 83.8 & 84.0 & 66.7 \\ % √
 Panoptic-PolarNet \cite{zhou_panoptic_polarnet_2021}& 63.6 & 67.1 & 84.3 & 75.1& 59.0 & 84.3 & 69.8 & 71.3 & 84.2& 83.9&  67.0 \\ % √
 Panoptic PH-Net \cite{li2022panoptic}& \textbf{80.1} & \textbf{82.8} & \textbf{91.1} & \textbf{87.6} & \textbf{82.1} & \text{93.0} & 88.1 & \textbf{76.6} & \textbf{87.9} & \textbf{86.6} & \textbf{80.2}\\
 \hline
 \textbf{LCPS (Baseline)} & 72.8 & 76.3 & 88.6 & 81.7 & 72.4 & 90.2 & 80.0 & 73.5 & 86.1 & 84.6 & 74.8 \\
 \textbf{LCPS (Full)}& 79.5 & 82.3 & 90.3 & \textbf{87.7} & 81.7 & 92.2 & \textbf{88.6} & 75.9 & 87.3 & 86.3 & 78.9 \\
\bottomrule %添加表格底部粗线
\end{tabular}
}
\setlength{\abovecaptionskip}{2pt}
\caption{3D panoptic segmentation results on NuScenes test set. Our result is compared with other methods without test-time augmentation and ensemble operations.}
\label{nusc_test}
\end{table*}

In this section, we evaluate our proposed LiDAR-Camera Panoptic Segmentation network on NuScenes \cite{fong2022panoptic} and SemanticKITTI \cite{behley2019SemanticKITTI} dataset, making comparisons with recent state-of-the-art methods. 

\subsection{Datasets and Evaluation Metric} \label{sec:datasets_and_metrics}

\noindent \textbf{NuScenes} is a large-scale multi-modal dataset for autonomous driving. It contains a 32-beam LiDAR, 5 Radars, 6 RGB cameras and maps, covering 1000 real-world driving scenes of 4 locations in Boston and Singapore. There are 850 annotated scenes for training and 150 for testing. The panoptic annotations contain 10 \textit{Thing} classes, 6 \textit{Stuff} classes and 1 class for noisy labels. 

\noindent \textbf{SemanticKITTI} is a pioneering outdoor dataset presenting the panoptic segmentation tasks on LiDAR data \cite{behley2019SemanticKITTI, behley2021ijrr, geiger2012cvpr}. It provides a 64-beam LiDAR sensor and two front-view cameras. The dataset contains 8 \textit{Thing} classes and 11 \textit{Stuff} classes, consisting of 19130 frames for training, 4071 for validation, and 20351 for testing.

\noindent \textbf{Evaluation Metrics.} 
We assess the panoptic segmentation via panoptic quality ($\text{PQ}$), segmentation quality ($\text{SQ}$), and recognition quality (\text{RQ}) \cite{fong2022panoptic}. Metrics with superscripts $\text{th}$ and $\text{st}$ (\textit{e.g., $PQ^\text{th}$}) represent \textit{Thing} or \textit{Stuff} classes performance respectively. Meanwhile, we also provide semantic segmentation metrics ($\text{mIoU}$) \cite{behley2019SemanticKITTI}.

\subsection{Implementation Details} \label{sec:implementation}

\noindent \textbf{Backbone Network.} Cylinder3D \cite{zhu2020cylindrical, DBLP:conf/cvpr/Hong0Z0L21} is adopted as our backbone network in Figure \ref{fig2:overview} due to its reliable LiDAR perception ability for cylinder voxel representation. As for NuScenes, the entire point cloud is divided into $480 \times 360 \times 32$ voxels for $[-100\text{m} \sim 100\text{m}, 0 \sim 2\pi, -5 \sim 3\text{m}]$ polarized volume of the scenery. As for SemanticKITTI, we only change the perception distance from $100\text{m}$ to $60\text{m}$.

\noindent \textbf{Settings and Hyper-parameters.}  
Following common practice \cite{li2022panoptic,xu_2022}, we apply random flip augmentation along the $y$-axis for the point cloud and images accordingly, and random rotation for the point cloud only. These LiDAR augmentations are adopted after precomputing the point-pixel alignment. The performance gains from data augmentations are already included in LiDAR-only baseline results for fair comparisons, as shown in the first line of Table \ref{ablation}. 
We train our model for 120 epochs with a batch size of $2$, using Adam optimizer \cite{DBLP:journals/corr/KingmaB14}. The initial learning rate is 0.004 and will be reduced to $0.0004$ after 100 epochs. For SARA described in Section \ref{sec:SARA}, the filtering parameter $\tau$ is set to 0.7. During inference, all operations are performed in BEV grids, where centers are picked from a dynamic heatmap using non-maximum-suppression with a kernel size of 5 and a value threshold of 0.1. Other setting details are described in the Appendix.

\noindent \subsection{Main Results} \label{sec:main_results}

In this section, we make extensive comparisons with other state-of-the-art methods and our LiDAR-only baseline. Specifically, the baseline network excludes the image branch, feature fusion module, and FOG in Figure \ref{fig2:overview}.

\noindent \textbf{Results on NuScenes.}
In Table \ref{nusc_val}, our approach outperforms the best Panoptic-PHNet \cite{li2022panoptic} by a margin of $5.1 \%$ $\text{PQ}$  ($79.8\%$ vs. $74.7\%$) in validation set.
Primarily, we achieve a large gain of $4.3\%$ $\text{RQ}$ and $7.1\%$ $\text{RQ}^{\text{th}}$, which mainly increases the overall accuracy. Compared with the LiDAR-only baseline, our methods show a significant improvement of $6.9\%$ $\text{PQ}$ in total, demonstrating the effectiveness of our LiDAR-Camera fusion strategy. As for the test set, we also achieve comparable SOTA results with Panoptic-PHNet \cite{li2022panoptic} without using test-time augmentation and $6.7 \%$ $\text{PQ}$ increase compared with our LiDAR-only baseline.

\begin{figure}[!htbp]
 \centering
 \includegraphics[width=0.5\textwidth]{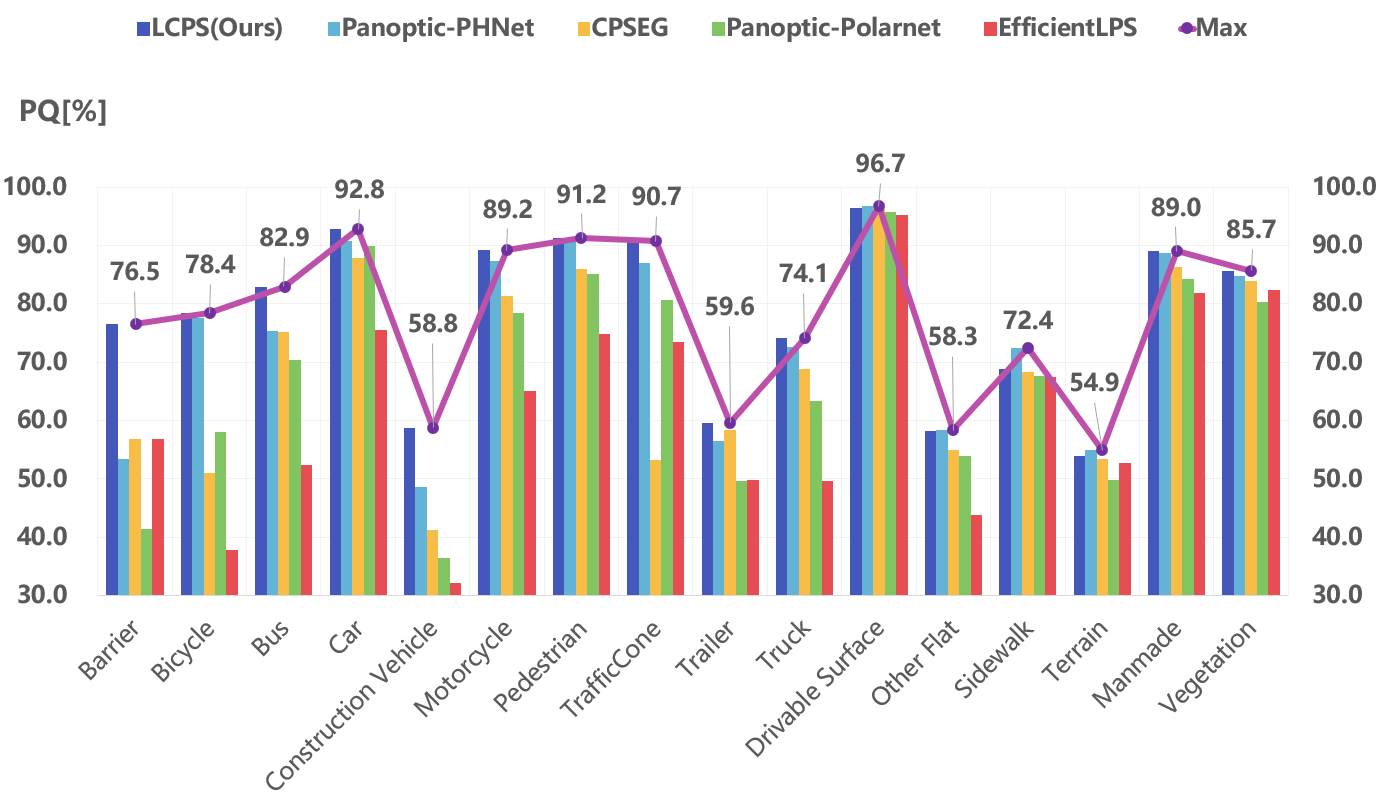} % Reduce the figure size so that it is slightly narrower than the column.
 \caption{Class-wise PQ\% results on NuScenes validation set.}
 
 \label{fig4:classwise_pq}
 \end{figure}

Evidence from the class-wise comparison on NuScenes validation set also consolidates the effectiveness of our fusion strategy. Figure \ref{fig4:classwise_pq} shows that an overall improvement among various \textit{Thing} and \textit{Stuff} categories can be witnessed. Specifically, for \textit{Thing} objects like \textit{bicycle}, \textit{bus}, \textit{construction vehicle}, \textit{motorcycle}, and \textit{traffic cone}, our method outperforms Panoptic-PHNet by a large margin ($9.3 \%$ on average for 5 \textit{Thing} classes), which demonstrates the ability of our approach to distinguish the sparse, distant and rare objects by taking advantages from image features.

\noindent \textbf{Results on SemanticKITTI.}
Here, we list the comparison results of the SemanticKITTI validation set in Table \ref{kitti_val}. Since SemanticKITTI has only two cameras in the front view, fewer points can be matched with image features compared with NuScenes, thus increasing the difficulty of LiDAR-Camera fusion. Nevertheless, we discover an increase of $3.3\%$ $\text{PQ}$ over the LiDAR-only baseline, demonstrating the robustness and effectiveness of our fusion strategy.

\begin{table}[!htbp]
\scriptsize
\centering
\resizebox{0.45\textwidth}{!}{
\begin{tabular}{l|cccc|c} %需要6列
\toprule %添加表格头部粗线
Method & PQ & PQ$^\dagger$ & SQ & RQ & mIoU\\
 %有n个&，就表示该行有n+1列
\hline %绘制一条水平横线
PanopticTrackNet \cite{hurtado2020mopt}& 40.0 & - & 73.0 & 48.3 & 53.8\\% √
DS-Net \cite{DBLP:conf/cvpr/Hong0Z0L21} & 57.7 & 63.4 & 77.6 & 68.0 & 63.5\\ % √
Panoptic-PolarNet \cite{zhou_panoptic_polarnet_2021} & 59.1 & 64.1 & 78.3 & 70.2 & 64.5\\ % √
EfficientLPS \cite{sirohi2021efficientlps}& 59.2 & 65.1 & 75.0 & 69.8 & 64.9\\ % √
Panoptic PH-Net \cite{li2022panoptic}& 61.7 & - & - & - & 65.7\\ % √
GP-S3Net \cite{razani_gp-s3net_2021}& 63.3 & 71.5 & 81.4 & 75.9 & 73.0\\
PUPS \cite{Su_2023} & \textbf{64.4} & \textbf{68.6} & \textbf{81.5} & \textbf{74.1} & -\\
 \hline
 \textbf{LCPS (Baseline)} & 55.7 & 65.2 & 74.0 & 65.8 & 61.1  \\
 \textbf{LCPS (Full)} & 59.0 & 68.8 & 79.8 & 68.9 & 63.2 \\
          
\bottomrule %添加表格底部粗线
\end{tabular}
}
\setlength{\abovecaptionskip}{0.05cm}
\caption{3D panoptic segmentation results on SemanticKITTI validation set.}
\label{kitti_val}
\end{table}

 \begin{figure*}[!htbp]
 \centering
 \includegraphics[width=1.0\textwidth]{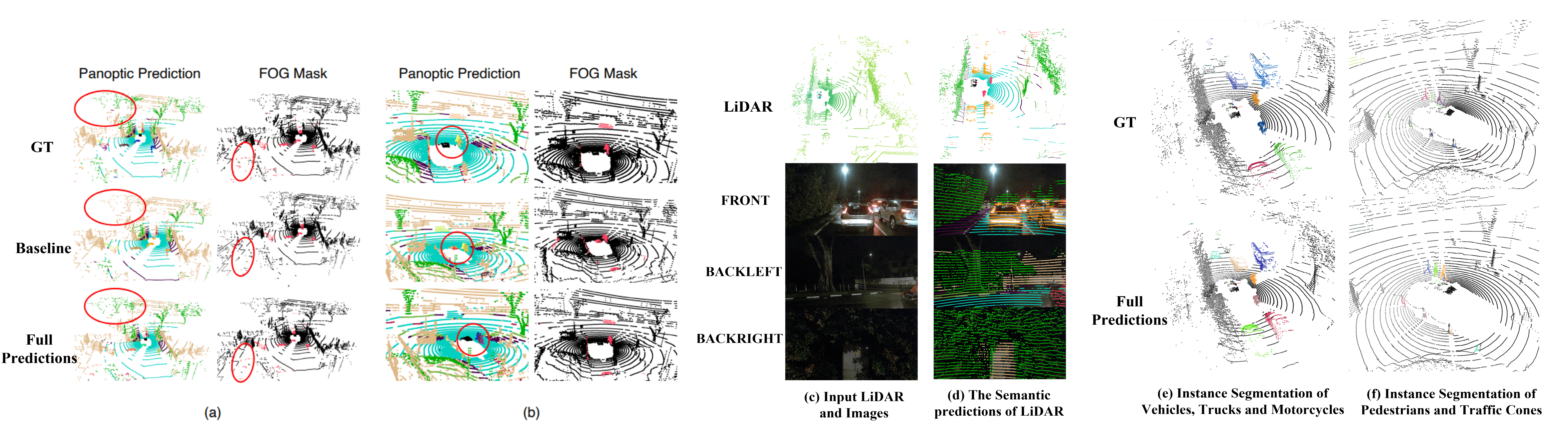}
 \caption{The overview of qualitative results from NuScenes validation set. (a) and (b) are visualization comparisons among the ground-truth (denoted as GT), the baseline predictions (Baseline), and full LCPS predictions (Full). Red circles emphasize the notable differences. We find that various \textit{Thing} and \textit{Stuff} objects can be predicted more accurately. (c) and (d) demonstrate semantic segmentation quality at nighttime. (e) and (f) verify the robust instance segmentation ability of our network. Better zoomed in.}
 \label{fig5:MainQua}
 \vspace{-1.2em}
 \end{figure*}

\begin{table}[!htbp]
\tiny
\centering
\resizebox{0.4\textwidth}{!}{
\begin{tabular}{cc|cc|c|c} %需要5列
\toprule %添加表格头部粗线
ACPA & SARA & PVP & SC & FOG & PQ\\
 %有n个&，就表示该行有n+1列
\hline %绘制一条水平横线
            &               &               &            &        & 72.9 \\
\checkmark  &               &               & \checkmark &        & 76.8 \\
\checkmark  &               & \checkmark    &            &        & 77.5 \\
\checkmark  & \checkmark    & \checkmark    &            &        & 79.2 \\
\hline
\checkmark  & \checkmark    & \checkmark    &            & \checkmark & 79.8 \\

\bottomrule %添加表格底部粗线
\end{tabular}
}
\setlength{\abovecaptionskip}{0.05cm}
\caption{Ablation study on NuScenes validation set. The SC represents Simple Concatenation compared to PVP.}
\label{ablation}
\vspace{-1.2em}
\end{table}

\begin{figure}[!htbp]
 \centering
 \includegraphics[width=0.5\textwidth]{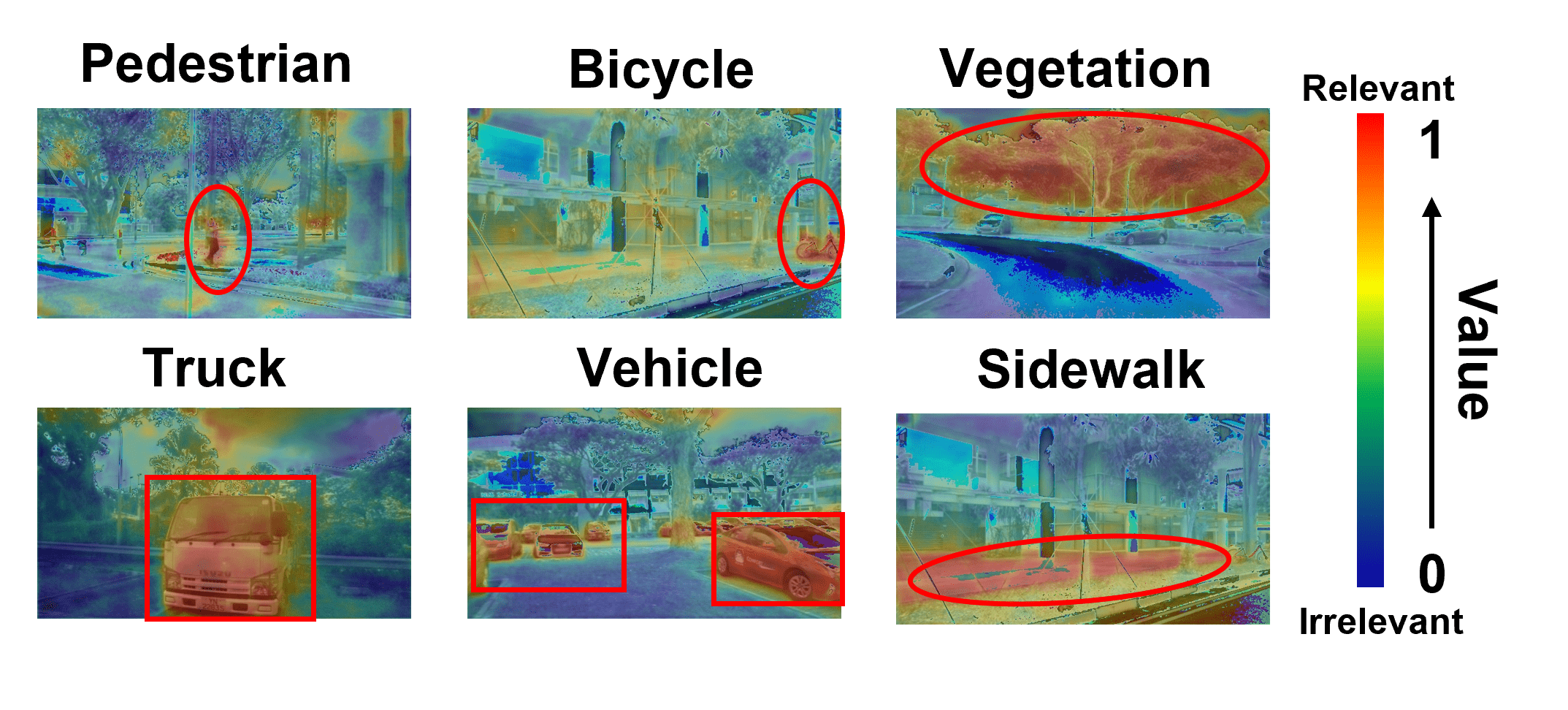} % Reduce the figure size so that it is slightly narrower than the column.
 \caption{The visualization results of semantic-aware regions filtered by CAMs.}
 \label{fig:cam}
 \vspace{-1.2em}
\end{figure}

\subsection{Ablation Study} \label{sec{ablation}}

\label{ablation_study}
To analyze the source of remarkable performance improvements, we conduct an ablation study on various components of our approach on NuScenes validation set. As depicted in Table \ref{ablation}, we divide the ablation study into the following three parts.

\noindent \textbf{Ablation on Fusion Modules.}
We separately verify the effectiveness of the Asynchronous Compensation Point Alignment (ACPA), Semantic-Aware Region Alignment (SARA), and Point-to-Voxel feature Propagation (PVP). It is observed that the ACPA with simple concatenation (SC) could bring an improvement of $3.9\%$ $\text{PQ}$ (contrasting line 1 and line 2) and an improvement of $4.6 \%$ $\text{PQ}$ with PVP module (contrasting line 1 and line 3). Another $1.7\%$ $\text{PQ}$ gain can be achieved combined with SARA (contrasting line 3 and line 4). It verifies our designs on geometry-consistent and semantic-aware LiDAR-Camera fusion strategy. 

\noindent \textbf{Ablation on FOG Mask.}
We test the influence of the FOG mask and observe the improvement of $0.6 \%$ $\text{PQ}$ (contrasting line 4 and line 5). It suggests that FOG Mask may bring additional supervision to the backbone network and further augments the semantic prediction in post-processing grouping.

\subsection{Qualitative results and the Discussion} \label{sec:qulitative}

\noindent \textbf{Visualization of Panoptic Predictions.}
In Figure \ref{fig5:MainQua}, we evaluate our visual predictions compared among ground-truth (GT), baseline, and full network (Full Predictions). The following observation can be made: 1) Our architecture achieves effective semantic and instance segmentation among challenging scenarios, like crowds of pedestrians and vehicles (see Figure \ref{fig5:MainQua} (a)(b)(e)(f)); 2) Our LiDAR-Camera Fusion strategies can achieve robust segmentation quality at nighttime with the complementary information from surrounding cameras (see Figure \ref{fig5:MainQua} (c)(d)); 3) FOG can help filter confusing points and noise points, making segmentation quality more robust (see Figure \ref{fig5:MainQua} (a)(b)).

\noindent \textbf{Visualization of Class Activation Maps.} We further verify the quality of generated Class Activation Maps (CAMs) in Figure \ref{fig:cam}, which constitute the semantic-aware regions in images. The red color illustrates higher semantic correlations, while the blue color refers to lower ones. It demonstrates that our SARA module generates highly correlated alignment among various categories, effectively extending the one-to-one mapping to semantic-aware one-to-many relations.

%------------------------------------------------------------------------
\section{The Conclusion} \label{sec:conclusion}

In this paper, we are the first to propose the geometry consistent and semantic-aware LiDAR-Camera Panoptic Network. As a new paradigm, we effectively exploit complementary information from LiDAR-Camera sensors and make essential efforts to overcome asynchronous and utilization problems via Asynchronous Compensation Point Alignment (ACPA), Semantic-Aware Region Alignment (SARA), Point-to-Voxel feature Propagation (PVP), and Foreground Object selection Gate (FOG) mask. These modules enhance the overall discriminability and performance. We hope that our thought-invoking multi-modal fusion practice can benefit future research.

\hspace*{\fill}

% \section{Acknowledgement} 
\noindent \textbf{Acknowledgement}
The research is supported by National Natural Science Foundation of China (No.62222602, No.61972157 and No.72192821), Shanghai Municipal Science and Technology Major Project (2021SHZDZX0102), Shanghai Science and Technology Commission (21511101200), Shanghai Sailing Program (22YF1420300 and 23YF1410500), Natural Science Foundation of Chongqing (No.CSTB2023NSCQ-JQX0007), CCF-Tencent Open Research Fund (RAGR20220121), Young Elite Scientists Sponsorship Program by CAST (2022QNRC001) and CAAI-Huawei MindSpore Open Fund.

{\small
\bibliographystyle{ieee_fullname}
\bibliography{egbib}

\begin{thebibliography}{10}\itemsep=-1pt

\bibitem{bai2022transfusion}
Xuyang Bai, Zeyu Hu, Xinge Zhu, Qingqiu Huang, Yilun Chen, Hongbo Fu, and
  Chiew-Lan Tai.
\newblock Transfusion: Robust lidar-camera fusion for 3d object detection with
  transformers.
\newblock In {\em Proceedings of the IEEE/CVF Conference on Computer Vision and
  Pattern Recognition}, pages 1090--1099, 2022.

\bibitem{behley2021ijrr}
J. Behley, M. Garbade, A. Milioto, J. Quenzel, S. Behnke, J. Gall, and C.
  Stachniss.
\newblock {Towards 3D LiDAR-based semantic scene understanding of 3D point
  cloud sequences: The SemanticKITTI Dataset}.
\newblock {\em The International Journal on Robotics Research},
  40(8-9):959--967, 2021.

\bibitem{behley2019SemanticKITTI}
Jens Behley, Martin Garbade, Andres Milioto, Jan Quenzel, Sven Behnke, Cyrill
  Stachniss, and Jurgen Gall.
\newblock Semantickitti: A dataset for semantic scene understanding of lidar
  sequences.
\newblock In {\em Proceedings of the IEEE/CVF International Conference on
  Computer Vision}, pages 9297--9307, 2019.

\bibitem{berman2018lovasz}
Maxim Berman, Amal~Rannen Triki, and Matthew~B Blaschko.
\newblock The lov{\'a}sz-softmax loss: A tractable surrogate for the
  optimization of the intersection-over-union measure in neural networks.
\newblock In {\em Proceedings of the IEEE conference on computer vision and
  pattern recognition}, pages 4413--4421, 2018.

\bibitem{chen2017deeplab}
Liang-Chieh Chen, George Papandreou, Iasonas Kokkinos, Kevin Murphy, and Alan~L
  Yuille.
\newblock Deeplab: Semantic image segmentation with deep convolutional nets,
  atrous convolution, and fully connected crfs.
\newblock {\em IEEE transactions on pattern analysis and machine intelligence},
  40(4):834--848, 2017.

\bibitem{Cheng_2020}
Bowen Cheng, Maxwell~D. Collins, Yukun Zhu, Ting Liu, Thomas~S. Huang, Hartwig
  Adam, and Liang-Chieh Chen.
\newblock Panoptic-{DeepLab}: A simple, strong, and fast baseline for bottom-up
  panoptic segmentation.
\newblock In {\em 2020 {IEEE}/{CVF} Conference on Computer Vision and Pattern
  Recognition ({CVPR})}. {IEEE}, jun 2020.

\bibitem{fong2022panoptic}
Whye~Kit Fong, Rohit Mohan, Juana~Valeria Hurtado, Lubing Zhou, Holger Caesar,
  Oscar Beijbom, and Abhinav Valada.
\newblock Panoptic nuscenes: A large-scale benchmark for lidar panoptic
  segmentation and tracking.
\newblock {\em IEEE Robotics and Automation Letters}, 7(2):3795--3802, 2022.

\bibitem{geiger2012cvpr}
A. Geiger, P. Lenz, and R. Urtasun.
\newblock {Are we ready for Autonomous Driving? The KITTI Vision Benchmark
  Suite}.
\newblock In {\em Proc.~of the IEEE Conf.~on Computer Vision and Pattern
  Recognition (CVPR)}, pages 3354--3361, 2012.

\bibitem{he2017mask}
Kaiming He, Georgia Gkioxari, Piotr Doll{\'a}r, and Ross Girshick.
\newblock Mask r-cnn.
\newblock In {\em Proceedings of the IEEE international conference on computer
  vision}, pages 2961--2969, 2017.

\bibitem{he2015spatial}
Kaiming He, Xiangyu Zhang, Shaoqing Ren, and Jian Sun.
\newblock Spatial pyramid pooling in deep convolutional networks for visual
  recognition.
\newblock {\em IEEE transactions on pattern analysis and machine intelligence},
  37(9):1904--1916, 2015.

\bibitem{DBLP:conf/cvpr/Hong0Z0L21}
Fangzhou Hong, Hui Zhou, Xinge Zhu, Hongsheng Li, and Ziwei Liu.
\newblock Lidar-based panoptic segmentation via dynamic shifting network.
\newblock In {\em {IEEE} Conference on Computer Vision and Pattern Recognition,
  {CVPR} 2021, virtual, June 19-25, 2021}, pages 13090--13099. Computer Vision
  Foundation / {IEEE}, 2021.

\bibitem{Huang_2020}
Tengteng Huang, Zhe Liu, Xiwu Chen, and Xiang Bai.
\newblock {EPNet}: Enhancing point features with image semantics for 3d object
  detection.
\newblock In {\em Computer Vision {\textendash} {ECCV} 2020}, pages 35--52.
  Springer International Publishing, 2020.

\bibitem{hurtado2020mopt}
Juana~Valeria Hurtado, Rohit Mohan, Wolfram Burgard, and Abhinav Valada.
\newblock Mopt: Multi-object panoptic tracking.
\newblock {\em The IEEE Conference on Computer Vision and Pattern Recognition
  (CVPR) Workshop on Scalability in Autonomous Driving}, 2020.

\bibitem{DBLP:journals/corr/KingmaB14}
Diederik~P. Kingma and Jimmy Ba.
\newblock Adam: {A} method for stochastic optimization.
\newblock In Yoshua Bengio and Yann LeCun, editors, {\em 3rd International
  Conference on Learning Representations, {ICLR} 2015, San Diego, CA, USA, May
  7-9, 2015, Conference Track Proceedings}, 2015.

\bibitem{kirillov2019panoptic}
Alexander Kirillov, Kaiming He, Ross Girshick, Carsten Rother, and Piotr
  Doll{\'a}r.
\newblock Panoptic segmentation.
\newblock In {\em Proceedings of the IEEE/CVF Conference on Computer Vision and
  Pattern Recognition}, pages 9404--9413, 2019.

\bibitem{Ku_2018}
Jason Ku, Melissa Mozifian, Jungwook Lee, Ali Harakeh, and Steven~L. Waslander.
\newblock Joint 3d proposal generation and object detection from view
  aggregation.
\newblock In {\em 2018 {IEEE}/{RSJ} International Conference on Intelligent
  Robots and Systems ({IROS})}. {IEEE}, oct 2018.

\bibitem{Li_2022}
Enxu Li, Ryan Razani, Yixuan Xu, and Bingbing Liu.
\newblock {SMAC}-seg: {LiDAR} panoptic segmentation via sparse
  multi-directional attention clustering.
\newblock In {\em 2022 International Conference on Robotics and Automation
  ({ICRA})}. {IEEE}, may 2022.

\bibitem{Li_2023}
Enxu Li, Ryan Razani, Yixuan Xu, and Bingbing Liu.
\newblock {CPSeg}: Cluster-free panoptic segmentation of 3d {LiDAR} point
  clouds.
\newblock In {\em 2023 {IEEE} International Conference on Robotics and
  Automation ({ICRA})}. {IEEE}, may 2023.

\bibitem{li2022panoptic}
Jinke Li, Xiao He, Yang Wen, Yuan Gao, Xiaoqiang Cheng, and Dan Zhang.
\newblock Panoptic-phnet: Towards real-time and high-precision lidar panoptic
  segmentation via clustering pseudo heatmap.
\newblock In {\em Proceedings of the IEEE/CVF Conference on Computer Vision and
  Pattern Recognition}, pages 11809--11818, 2022.

\bibitem{li2022deepfusion}
Yingwei Li, Adams~Wei Yu, Tianjian Meng, Ben Caine, Jiquan Ngiam, Daiyi Peng,
  Junyang Shen, Yifeng Lu, Denny Zhou, Quoc~V Le, et~al.
\newblock Deepfusion: Lidar-camera deep fusion for multi-modal 3d object
  detection.
\newblock In {\em Proceedings of the IEEE/CVF Conference on Computer Vision and
  Pattern Recognition}, pages 17182--17191, 2022.

\bibitem{Liang_2018}
Ming Liang, Bin Yang, Shenlong Wang, and Raquel Urtasun.
\newblock Deep continuous fusion for multi-sensor 3d object detection.
\newblock In {\em Computer Vision {\textendash} {ECCV} 2018}, pages 663--678.
  Springer International Publishing, 2018.

\bibitem{lim2020learning}
Long~Ang Lim and Hacer~Yalim Keles.
\newblock Learning multi-scale features for foreground segmentation.
\newblock {\em Pattern Analysis and Applications}, 23(3):1369--1380, 2020.

\bibitem{liu2023bevfusion}
Zhijian Liu, Haotian Tang, Alexander Amini, Xinyu Yang, Huizi Mao, Daniela~L
  Rus, and Song Han.
\newblock Bevfusion: Multi-task multi-sensor fusion with unified bird's-eye
  view representation.
\newblock In {\em 2023 IEEE International Conference on Robotics and Automation
  (ICRA)}, pages 2774--2781. IEEE, 2023.

\bibitem{mcever2020pcams}
R~Austin McEver and BS Manjunath.
\newblock Pcams: Weakly supervised semantic segmentation using point
  supervision.
\newblock {\em arXiv preprint arXiv:2007.05615}, 2020.

\bibitem{milioto2020lidar}
Andres Milioto, Jens Behley, Chris McCool, and Cyrill Stachniss.
\newblock Lidar panoptic segmentation for autonomous driving.
\newblock In {\em 2020 IEEE/RSJ International Conference on Intelligent Robots
  and Systems (IROS)}, pages 8505--8512. IEEE, 2020.

\bibitem{prakash2021multi}
Aditya Prakash, Kashyap Chitta, and Andreas Geiger.
\newblock Multi-modal fusion transformer for end-to-end autonomous driving.
\newblock In {\em Proceedings of the IEEE/CVF Conference on Computer Vision and
  Pattern Recognition}, pages 7077--7087, 2021.

\bibitem{Qi_2018}
Charles~R. Qi, Wei Liu, Chenxia Wu, Hao Su, and Leonidas~J. Guibas.
\newblock Frustum {PointNets} for 3d object detection from {RGB}-d data.
\newblock In {\em 2018 {IEEE}/{CVF} Conference on Computer Vision and Pattern
  Recognition}. {IEEE}, jun 2018.

\bibitem{razani_gp-s3net_2021}
Ryan Razani, Ran Cheng, Enxu Li, Ehsan Taghavi, Yuan Ren, and Liu Bingbing.
\newblock {GP}-{S3Net}: {Graph}-based panoptic sparse semantic segmentation
  network.
\newblock In {\em Proceedings of the {IEEE}/{CVF} {International} {Conference}
  on {Computer} {Vision}}, pages 16076--16085, 2021.

\bibitem{Sindagi_2019}
Vishwanath~A. Sindagi, Yin Zhou, and Oncel Tuzel.
\newblock {MVX}-net: Multimodal {VoxelNet} for 3d object detection.
\newblock In {\em 2019 International Conference on Robotics and Automation
  ({ICRA})}. {IEEE}, may 2019.

\bibitem{sirohi2021efficientlps}
Kshitij Sirohi, Rohit Mohan, Daniel B{\"u}scher, Wolfram Burgard, and Abhinav
  Valada.
\newblock Efficientlps: Efficient lidar panoptic segmentation.
\newblock {\em IEEE Transactions on Robotics}, 2021.

\bibitem{Su_2023}
Shihao Su, Jianyun Xu, Huanyu Wang, Zhenwei Miao, Xin Zhan, Dayang Hao, and Xi
  Li.
\newblock {PUPS}: Point cloud unified panoptic segmentation.
\newblock {\em Proceedings of the {AAAI} Conference on Artificial
  Intelligence}, 37(2):2339--2347, jun 2023.

\bibitem{vaswani2017attention}
Ashish Vaswani, Noam Shazeer, Niki Parmar, Jakob Uszkoreit, Llion Jones,
  Aidan~N Gomez, {\L}ukasz Kaiser, and Illia Polosukhin.
\newblock Attention is all you need.
\newblock {\em Advances in neural information processing systems}, 30, 2017.

\bibitem{Vora_2020}
Sourabh Vora, Alex~H. Lang, Bassam Helou, and Oscar Beijbom.
\newblock {PointPainting}: Sequential fusion for 3d object detection.
\newblock In {\em 2020 {IEEE}/{CVF} Conference on Computer Vision and Pattern
  Recognition ({CVPR})}. {IEEE}, jun 2020.

\bibitem{Wang_2021}
Chunwei Wang, Chao Ma, Ming Zhu, and Xiaokang Yang.
\newblock {PointAugmenting}: Cross-modal augmentation for 3d object detection.
\newblock In {\em 2021 {IEEE}/{CVF} Conference on Computer Vision and Pattern
  Recognition ({CVPR})}. {IEEE}, jun 2021.

\bibitem{wang2021pointaugmenting}
Chunwei Wang, Chao Ma, Ming Zhu, and Xiaokang Yang.
\newblock Pointaugmenting: Cross-modal augmentation for 3d object detection.
\newblock In {\em Proceedings of the IEEE/CVF Conference on Computer Vision and
  Pattern Recognition}, pages 11794--11803, 2021.

\bibitem{Wang_2021_CVPR}
Haochen Wang, Xiaolong Jiang, Haibing Ren, Yao Hu, and Song Bai.
\newblock Swiftnet: Real-time video object segmentation.
\newblock In {\em Proceedings of the IEEE/CVF Conference on Computer Vision and
  Pattern Recognition (CVPR)}, pages 1296--1305, June 2021.

\bibitem{wang2021swiftnet}
Haochen Wang, Xiaolong Jiang, Haibing Ren, Yao Hu, and Song Bai.
\newblock Swiftnet: Real-time video object segmentation.
\newblock In {\em Proceedings of the IEEE/CVF Conference on Computer Vision and
  Pattern Recognition}, pages 1296--1305, 2021.

\bibitem{xu_2022}
Shuangjie Xu, Rui Wan, Maosheng Ye, Xiaoyi Zou, and Tongyi Cao.
\newblock Sparse cross-scale attention network for efficient {LiDAR} panoptic
  segmentation.
\newblock {\em Proceedings of the {AAAI} Conference on Artificial
  Intelligence}, 36(3):2920--2928, jun 2022.

\bibitem{Zhou_2016}
Bolei Zhou, Aditya Khosla, Agata Lapedriza, Aude Oliva, and Antonio Torralba.
\newblock Learning deep features for discriminative localization.
\newblock In {\em 2016 {IEEE} Conference on Computer Vision and Pattern
  Recognition ({CVPR})}. {IEEE}, jun 2016.

\bibitem{zhou2021probablistic}
Xingyi Zhou, Vladlen Koltun, and Philipp Kr{\"a}henb{\"u}hl.
\newblock Probabilistic two-stage detection.
\newblock In {\em arXiv preprint arXiv:2103.07461}, 2021.

\bibitem{zhou_panoptic_polarnet_2021}
Zixiang Zhou, Yang Zhang, and Hassan Foroosh.
\newblock Panoptic-polarnet: {Proposal}-free lidar point cloud panoptic
  segmentation.
\newblock In {\em Proceedings of the {IEEE}/{CVF} {Conference} on {Computer}
  {Vision} and {Pattern} {Recognition}}, pages 13194--13203, 2021.

\bibitem{zhu2020cylindrical}
Xinge Zhu, Hui Zhou, Tai Wang, Fangzhou Hong, Yuexin Ma, Wei Li, Hongsheng Li,
  and Dahua Lin.
\newblock Cylindrical and asymmetrical 3d convolution networks for lidar
  segmentation.
\newblock {\em arXiv preprint arXiv:2011.10033}, 2020.

\bibitem{zhuang2021perception}
Zhuangwei Zhuang, Rong Li, Kui Jia, Qicheng Wang, Yuanqing Li, and Mingkui Tan.
\newblock Perception-aware multi-sensor fusion for 3d lidar semantic
  segmentation.
\newblock In {\em Proceedings of the IEEE/CVF International Conference on
  Computer Vision}, pages 16280--16290, 2021.

\bibitem{Zhuang_2021}
Zhuangwei Zhuang, Rong Li, Kui Jia, Qicheng Wang, Yuanqing Li, and Mingkui Tan.
\newblock Perception-aware multi-sensor fusion for 3d {LiDAR} semantic
  segmentation.
\newblock In {\em 2021 {IEEE}/{CVF} International Conference on Computer Vision
  ({ICCV})}. {IEEE}, oct 2021.

\end{thebibliography}
}

%------------------------------------------------------------------------
\clearpage 
\appendix
\section*{Appendix}

\begin{subappendices}
\subsection*{A. Further Implementation Details}
In this section, we further elaborate on the implementation details of our LCPS.
Section \ref{sec:overview} in the main paper explains that the LiDAR branch consists of point and voxel streams. We employ three MLP layers for the point stream to extract point-level features with an output dimension of $64$, $128$, and $256$ channels. Following processing by ACPA and SARA, the fused point features are compressed to $16$ channels to match the voxel features. As for the voxel stream, the cylindrical encoder maps original point features (XYZ-axis, remission, reflections, etc) to $16$ dimensions. After PVP module, features from the voxel stream and point stream are merged into cylinders and then fed into Cylinder3D \cite{zhu2020cylindrical} backbone network. In Cylinder3D, four layers of down-sampling 3D convolutions with BatchNorm and ReLU activation are applied to the fused voxel features, transforming the channel numbers of voxel features to $32$, $64$, $128$, and $256$, respectively. Finally, the voxel feature dimension is compressed back to $128$ after pooling layers and remains at this dimension in the subsequent four up-sampling layers of Cylinder3D.

Regarding the image branch, images from multiple cameras are concatenated and scaled to 0.4 of the original size. The SwiftNet-18\cite{wang2021swiftnet} network comprises four pairs of down-sampling and up-sampling layers. The four down-sampling layers transform features maps to $64 \times 320 \times 180$, $128 \times 160 \times 90$, $256 \times\ 80 \times 45$, and $512 \times 40 \times 23$, respectively. Then, a multi-scale spatial pooling module \cite{he2015spatial} is utilized to compress the four feature maps to $128$ channels. Eventually, the other 4 up-sampling layers symmetrically up-sample the feature map to the input size for following geometric-consistent and semantic-aware alignment and fusion.

\subsection*{B. Further Discussion}

\begin{figure*}[!htbp]
 \centering
 \includegraphics[width=0.8\textwidth]{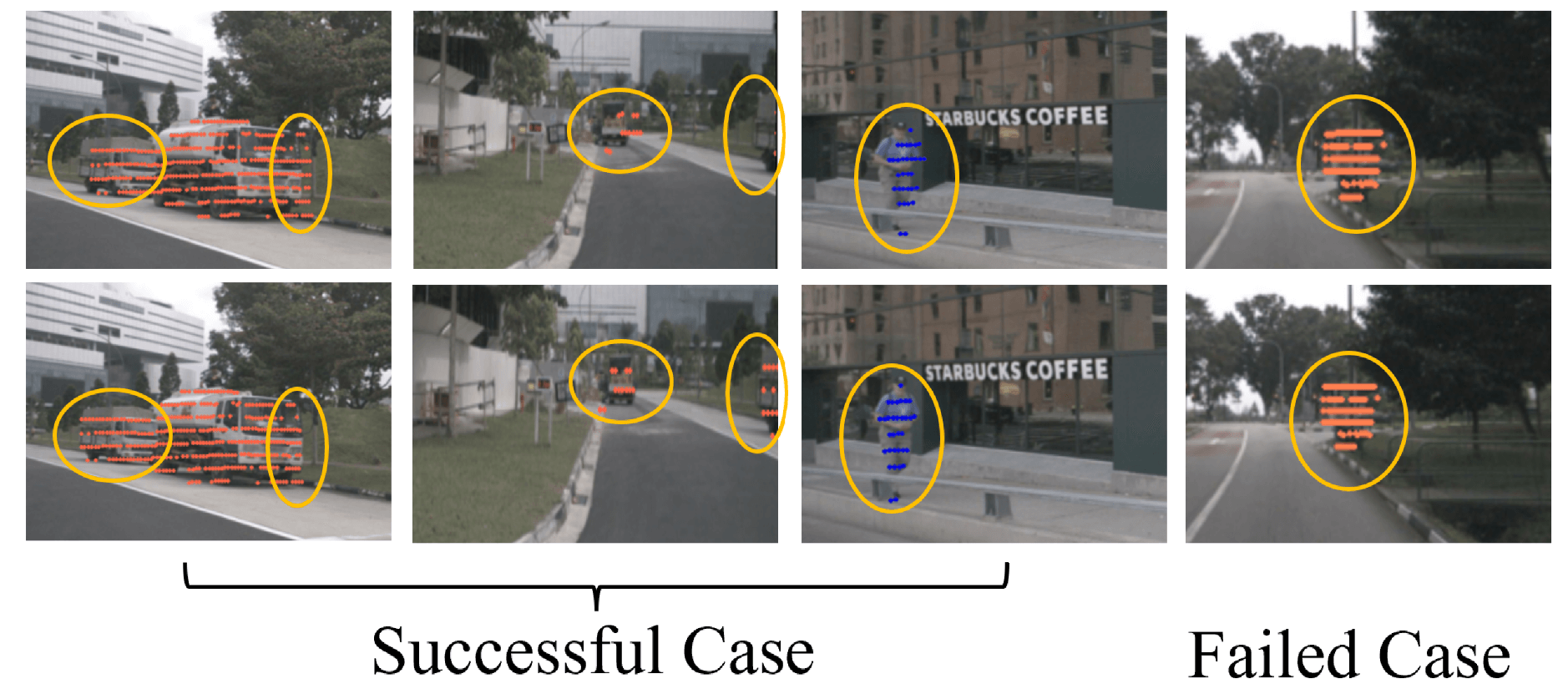} % Reduce the figure size so that it is slightly narrower than the column.
 \caption{Visual comparisons of asynchronous compensation. The first and the second lines are visualizations without or with asynchronous compensation respectively. The leftmost three columns demonstrate the effectiveness, especially for foreground objects of various sizes and geometric shapes. The last column specifies that the asynchronous compensation almost makes no difference when the time gap is small or at the front view.}
 \label{fig:proj_compare}
\end{figure*}

\noindent \textbf{Visual Ablations on Asynchronous Compensation.}
Here we further provide qualitative comparisons of asynchronous compensation in Figure \ref{fig:proj_compare}. The first and the second lines are visual results without or with asynchronous compensation respectively. The leftmost three columns demonstrate the effectiveness, especially for foreground objects of various sizes and distinct geometric shapes. For instance, the most apparent improvement is exhibited in the second leftmost column, where few points can be mapped to distant and marginal trucks, greatly enhancing the robustness of LiDAR-Camera fusion. We also demonstrate a typical failed case here to illustrate the limitation. When the ego-vehicle slows down, or objects come at the front or back view, it is possible that the asynchronous compensation almost
makes no difference because the time gap or the changes of view angles is small.

\noindent \textbf{Discussions on Time and Memory Cost.}
We compare the time and memory cost in Table \ref{tab:params} with other SOTA approaches if their projects are open-source or if such information is provided in their papers. Our LCPS full model is slightly slower than LiDAR-only methods (including our LiDAR-only baseline). Interestingly, adding image branches does not essentially drop the FPS since we choose lightweight ResNet-18 as the image backbone. The parameter number of our LiDAR-only baseline is high since we replace the backbone BEV U-Net in Panoptic-Polarnet \cite{zhou_panoptic_polarnet_2021} with Cylinder3D \cite{zhu2020cylindrical}. Similarly, the parameter number of DS-Net \cite{DBLP:conf/cvpr/Hong0Z0L21} is also above 50M since it adopts Cylinder3D as the backbone network too.

\begin{table}[htbp!]
\vspace{-0.3cm}
\centering
\tiny
\resizebox{0.5\textwidth}{!}{
\begin{tabular}{c|c|c}
\hline
        & FPS(Hz) & Params(M) \\\hline
DS-Net \cite{DBLP:conf/cvpr/Hong0Z0L21}                          & 3.2    & 56.5   \\
Panoptic-PolarNet \cite{zhou_panoptic_polarnet_2021}              & 11.6   & 13.8   \\
Panoptic-PHNet \cite{li2022panoptic}                 & 11.0   & -   \\ 
LCPS(LiDAR-Only)                                 & 8.6    & 65.9  \\
LCPS(Full)                                    & 8.3    & 77.7 \\
\hline
\end{tabular}
}
\setlength{\abovecaptionskip}{0.05cm}
\caption{Results of FPS and parameter scales on SemanticKITTI.}
\label{tab:params}
\end{table}

\noindent \textbf{More In-Depth Analysis on Metrics.}
We provide further in-depth analysis of our experimental results. It appears that some \textit{Stuff} metrics, e.g., $\text{PQ}^{\text{st}}$ and $\text{mIoU}$, are slightly lower than Panoptic-PHNet ($0.3\% $-$0.7 \%$) on NuScenes val. and test set. However, compared to the LiDAR-only baseline, our approach boosts the performance in terms of $\text{PQ}^{\text{st}}$, $\text{SQ}^{\text{st}}$, $\text{RQ}^{\text{st}}$, and $\text{mIoU}$ consistently. This phenomenon reflects that the improvement on \textit{Stuff} is not as noticeable as \textit{Thing} objects and further indicates that the LiDAR-Image fusion has more benefits on \textit{Thing} objects. The possible reason is that \textit{Thing} objects often have fewer points than \textit{Stuff}; thus, images may provide more crucial information for the former.

Our experimental results do not obviously surpass SOTA methods since our baseline is relatively weak. Our baseline project is built on the current highest open-source benchmark, Panoptic-Polarnet \cite{zhou_panoptic_polarnet_2021}, while other SOTA methods have not released their codes yet. We reproduce the Panoptic-Polarnet and get $67.7 \%$ PQ on NuScenes, and $55.7 \%$ PQ on SemanticKITTI. Then we further improve the NuScenes baseline to $72.9 \%$ using data augmentations as stated in Section \ref{sec:implementation}. Based on this baseline, our fusion strategy obtains $+6.9\%$, $+6.7\%$, and $+3.3 \% $ PQ improvement on NuScenes validation, NuScenes test, and SemanticKITTI validation set, reported in the main content. After submission, we improve our baseline on SemanticKITTI by using rare \textit{Stuff} copy-paste augmentation, demonstrating higher overall performances, as shown in Table \ref{tab:improved_kitti}. We provide the improved version here for additional reference and analysis. The fusion improvement on SemanticKITTI is lower than NuScenes since SemanticKITTI has only two front-view cameras; thus, the number of matched points is lower than NuScenes, as illustrated in Table \ref{tab:statistics}. Additionally, the improvement on a higher SemanticKITTI baseline is lower than the original one because heavy data augmentation (we add extra rare stuff augmentation in order to achieve a higher baseline) may diminish the benefits of LiDAR-Camera fusion, which is also reported in previous research on detection tasks like PointAugmenting \cite{wang2021pointaugmenting}.

As for mIoU, it is mainly evaluated for semantic segmentation, and we include it following previous research \cite{li2022panoptic, zhou_panoptic_polarnet_2021}. Better PQ simultaneously needs better semantic segmentation ability (mIoU) and instance segmentation quality. Therefore, a model of high mIoU may perform worse in terms of PQ. Our method can achieve comparable mIoU performance with SOTA Panoptic-PHNet in NuScenes, while worse in SemanticKitti due to the weak baseline issue. Besides, our fusion strategy consistently improves PQ and mIoU in both NuScenes and SemanticKitti datasets compared to the LiDAR-only baseline.
 
\begin{table}[htbp]
\vspace{-0.3cm}
\tiny
\centering
\resizebox{0.46\textwidth}{!}{
\begin{tabular}{c|c|c}
\hline
                        & NuScenes       &  SemanticKITTI      \\ \hline
Matched Points          &  17182         &   39780             \\
Total Points            &  34720         &   120387             \\
Percentage              &  52.2 \%       &   33.2 \%            \\ \hline
\end{tabular}}
\setlength{\abovecaptionskip}{0.05cm}
\caption{Statistics on the averaged number of points matched to images per frame.}
\label{tab:statistics}
\end{table}

\begin{table}[htbp]
\vspace{-0.3cm}
\centering
\tiny
\resizebox{0.5\textwidth}{!}{
\begin{tabular}{c|cc|cc}
\hline
\multirow{2}{*}{Methods}   &\multicolumn{2}{c|}{Improved (Val.)} &  \multicolumn{2}{c}{Improved (Test.)}   \\ \cline{2-5}
            &     PQ   &   mIoU  &  PQ  &   mIoU    \\ \hline
LCPS(LiDAR-Only)  &    60.6  &   66.8  &  57.8 &  62.0    \\
LCPS(Full) &    61.4  &   67.5  &  58.8 &  62.8   \\  \hline

\end{tabular}
}

\setlength{\abovecaptionskip}{0.05cm}
\caption{Results of the fusion methods on the improved baseline of SemanticKITTI.}
\label{tab:improved_kitti}
\end{table}

\noindent \textbf{Ablation Study on Perception Distance.}
As our backbone network adopts a cylindrical voxel representation, we need to set the perception distance of the scene volume, which is defined as the radial distance from the LiDAR sensor to objects or points. Setting the perception distance too close or too far is sub-optimal for training because a close distance setting may miss some small objects far away and diminish $\text{PQ}$ performance, while a far distance setting may involve more noise points and disturb training stability. 

In our experimentation on the NuScenes validation set (as shown in Table \ref{tab:distance}), we find that as the perception distance increases, the performance initially improves and then declines. The result shows that $\pm 100$ meters and $\pm 120$ meters yield the highest $\text{PQ}$ scores, while $\pm80$ meters produce the best $\text{mIoU}$. Intuitively, $\pm 80$ meters can be the valid distance at which the LiDAR sensor is able to accurately detect objects in NuScenes, while approximately $\pm 150$ meters is the farthest perception distance.
Based on these findings, we ultimately choose $\pm100$ meters as the moderate perception distance for NuScenes.

\begin{table*}[htbp!]
\tiny
\centering
\resizebox{1.0\textwidth}{!}{
% \scalebox{0.65}{
\begin{tabular}{l|cccccccccc} %需要11列
\toprule %添加表格头部粗线
Distance ($\pm$, [meters]) & 50 & 60 & 70 & 80 & 90 & 100 & 110 & 120 & 130 & NFV\\
 %有n个&，就表示该行有n+1列
\midrule %绘制一条水平横线
 PQ [\%]     & 70.6 & 72.2 & 72.5 & 72.8 & 72.8 & \textbf{72.9} & 72.2 & \textbf{72.9} & 72.3 & 70.5  \\ % √
 \hline
 mIoU [\%]   & 74.3 & 74.6 & 74.0 & \textbf{75.2} & 74.5 & 75.1 & 74.4 & 74.5 & 74.4 & 73.7       \\
\bottomrule %添加表格底部粗线
\end{tabular}
% }
}
\setlength{\abovecaptionskip}{0.05cm}
\caption{The ablation of perception distance on NuScenes validation set. The experiment is tuned on our LiDAR-only baseline network. NFV represents No Fixed Volume, which means we select the farthest point (usually $\ge 170\text{m}$) as the distance for each LiDAR scan rather than a fixed distance.}
\label{tab:distance}
\end{table*}

\begin{figure*}[!htbp]
 \centering
 \includegraphics[width=0.8\textwidth]{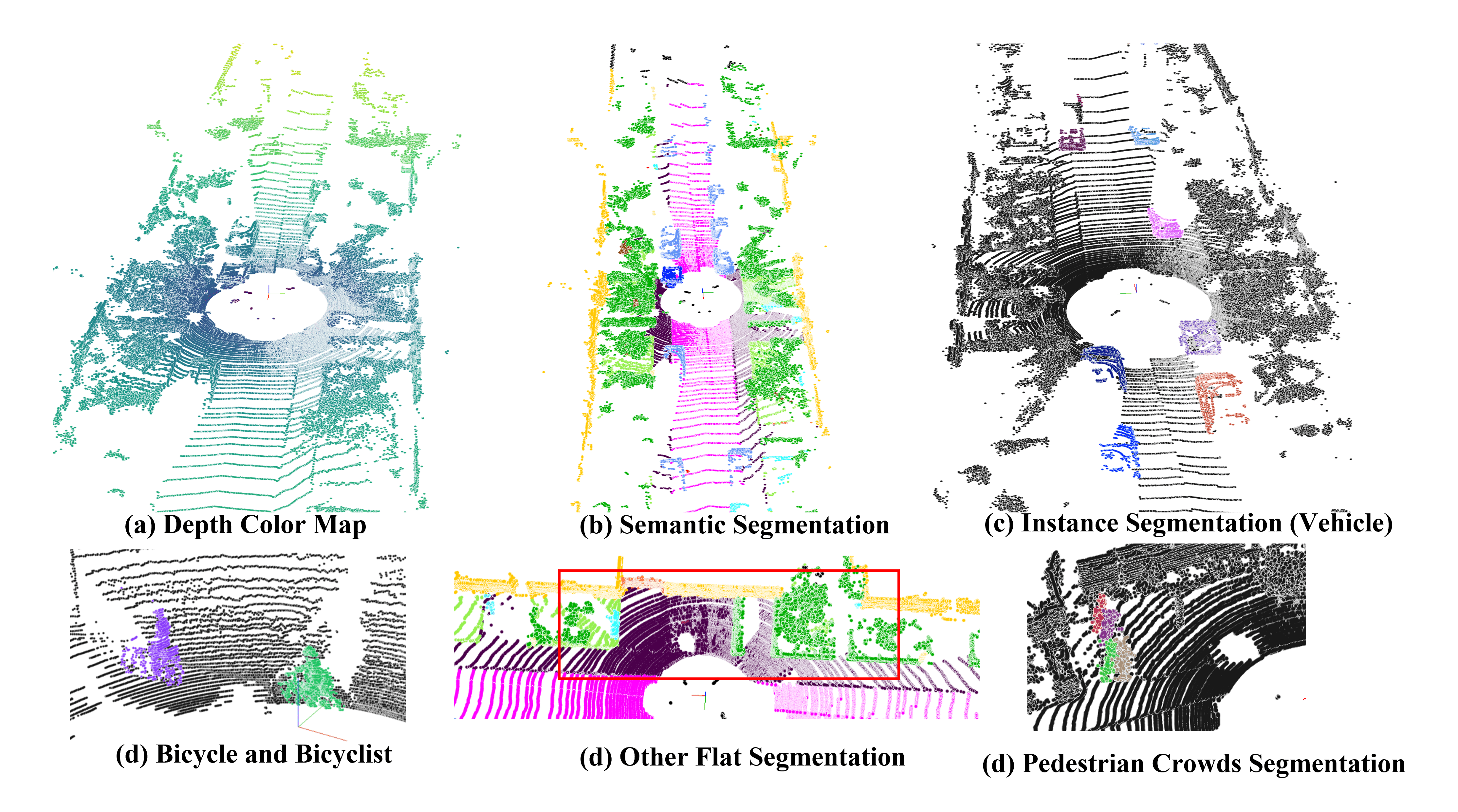} % Reduce the figure size so that it is slightly narrower than the column.
 \caption{Visualization results of SemanticKITTI validation set.}
 \label{fig:kitti}
 \end{figure*}

\begin{figure*}[!htbp]
 \centering
 \includegraphics[width=1.0\textwidth]{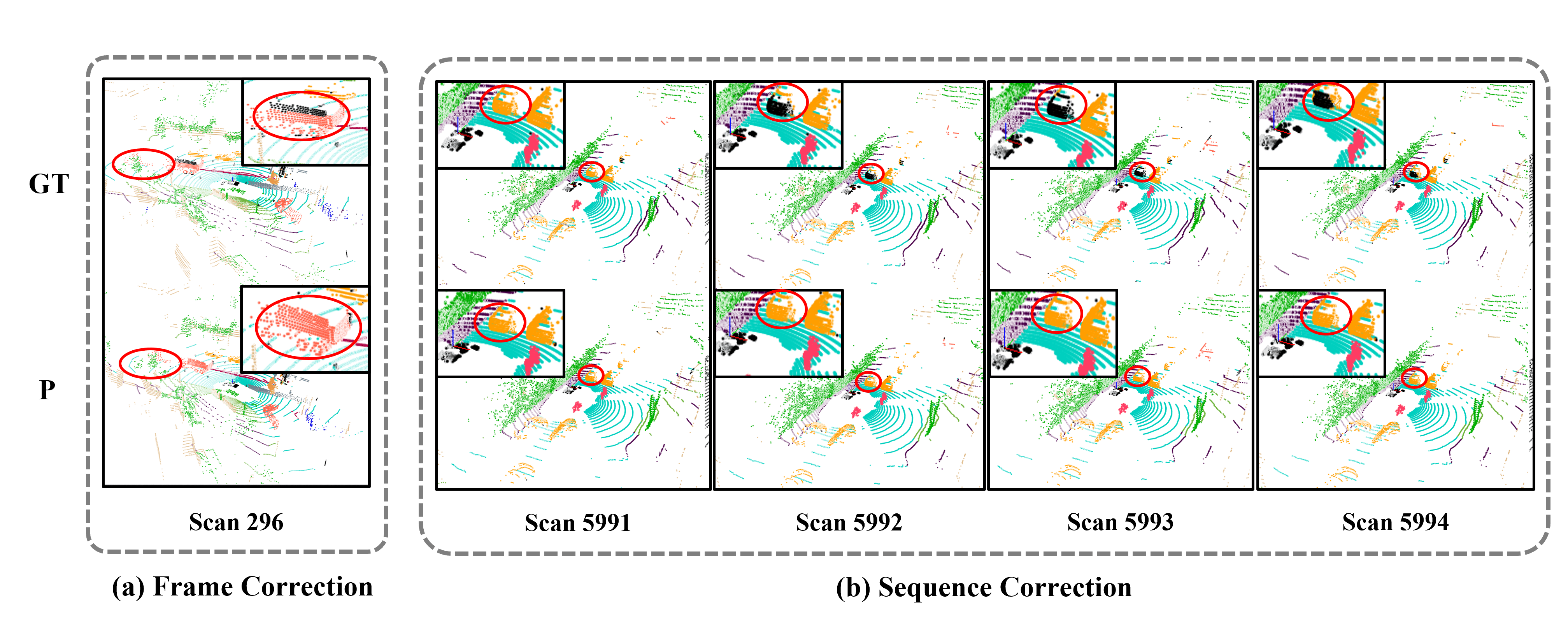} %
 \caption{The visualization of correction ability between ground-truth (GT, first line) and our model predictions (P, second line). The results are from the NuScenes validation set, where the scan number below represents the sample index and red circles highlight notable differences. (a) shows that our model can segment an intact segmentation of truck, even though the ground-truth labels overlook the top area. (b) demonstrates that our model can consistently segment an entire vehicle during a sequence of frames over time, while the ground-truth labels miss the back surface when the ego-car advances at a high speed.}
 
\label{fig:correction}

\end{figure*}

\noindent \textbf{Correction Ability.}
When we review the visualization results, one interesting observation is that our model appears to correct segmentation errors in the ground-truth labels. Due to the utilization of bounding boxes and semantic labels in rule-based scripts for automatically generating panoptic labels, errors in ground-truth labels are commonly observed. Hence, it is essential for our network to boost more generalizability and avoid overfitting with ground-truth labels during the training process.
 
 As illustrated in Figure \ref{fig:correction}, our LCPS network is capable of correcting ground-truth errors by preserving the intact shape of an instance. 
Figure \ref{fig:correction} (a) demonstrates that the predicted truck segmentation retains the top edge area since it is spatially and geometrically proximate to the truck segments below. Similarly, in a sequence of frames in Figure \ref{fig:correction} (b), the rule-based ground-truth label generation results in the absence of the back surface of the vehicle, as the ego-car advances at high speed. Nevertheless, our network still maintains consistency in predicted segmentation over time, which serves as compelling evidence that our network obtains robust feature representation of objects by leveraging LiDAR and image features, enabling it to correct false ground-truth labels.

\begin{figure*}[!htbp]
 \centering
 \includegraphics[width=1.0\textwidth]{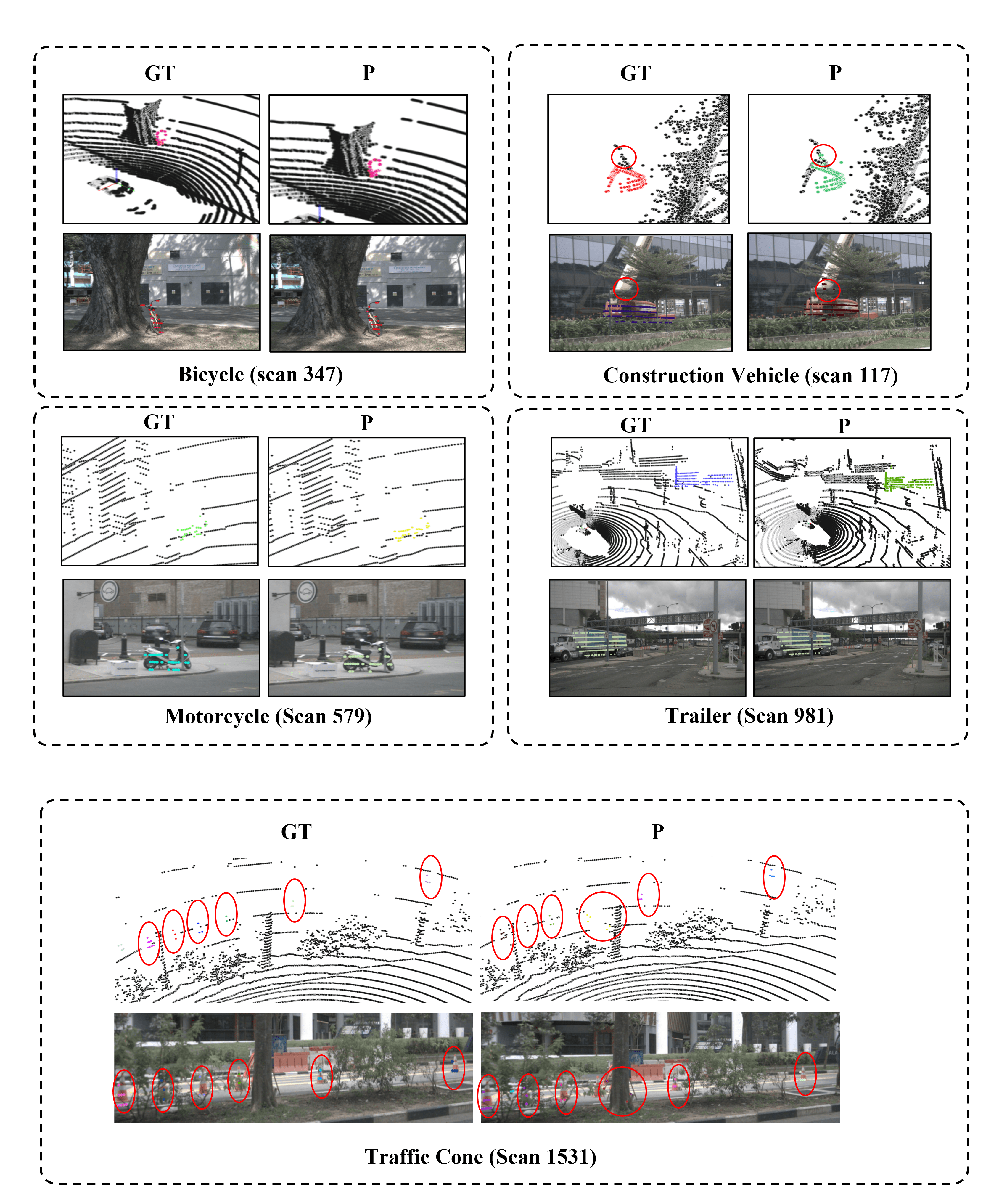} % Reduce the figure size so that it is slightly narrower than the column.
 \caption{Visualization results of foreground objects. GT represents ground-truth labels, while P represents predictions of our LCPS. Text such as "Back" and "FRONT\_LEFT" refers to the specific camera sensor. In this figure, the perspective projections of object segmentation are within the same image. Generally, our network achieves accurate segmentation results over these small and distant objects, such as \textit{bicycle} and \textit{motorcycle}, or rare objects like \textit{construction vehicle}.}
 \label{fig:single_view}
 \end{figure*}

\begin{figure*}[!htbp]
 \centering
 \includegraphics[width=1.0\textwidth]{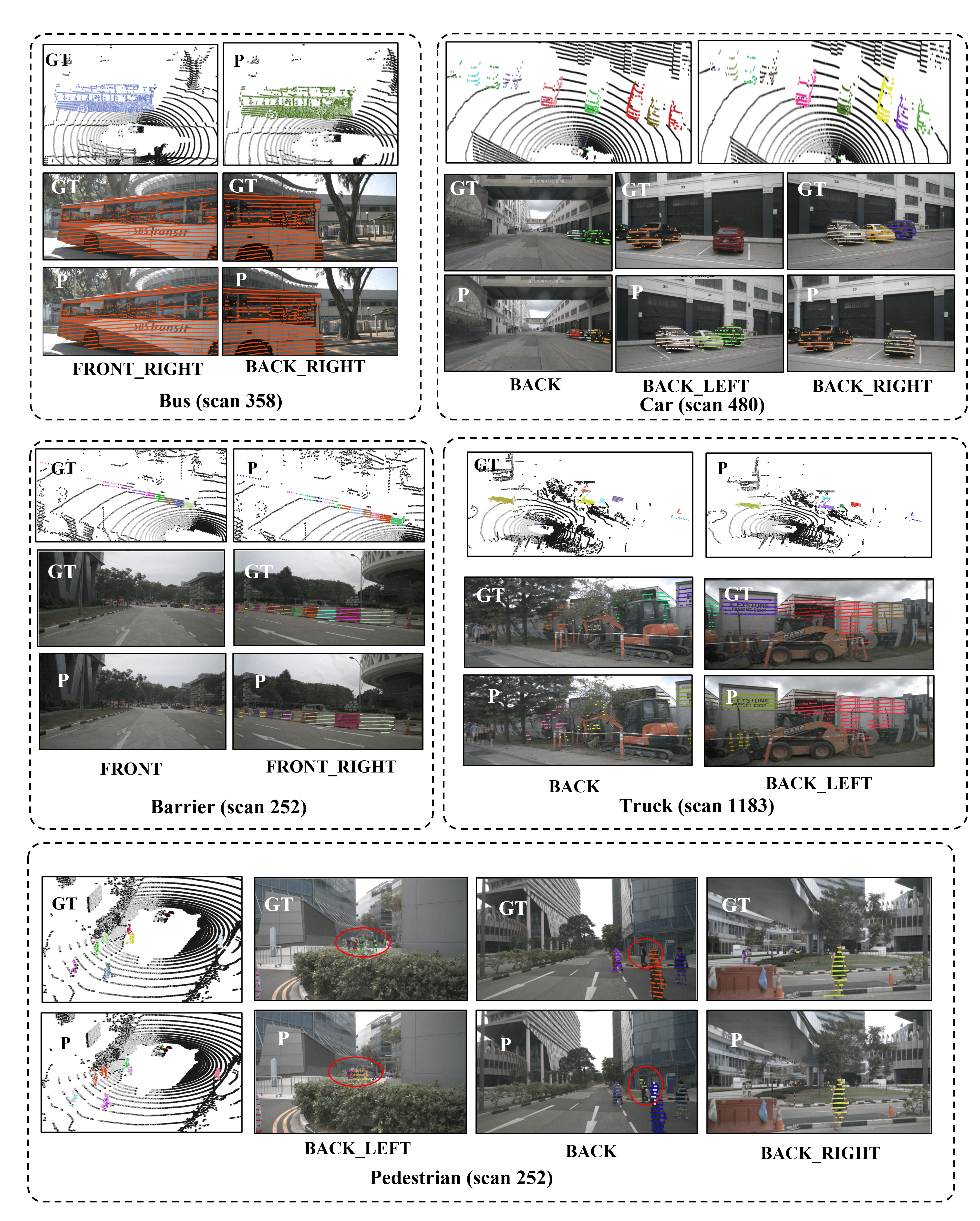} % Reduce the figure size so that it is slightly narrower than the column.
 \caption{Visualization results of foreground objects. GT represents ground-truth labels, while P represents predictions of our LCPS. Texts like "Back" and "FRONT\_LEFT" refer to the specific camera sensor. This figure shows that most objects of diverse types and spatial locations across images can be consistently identified.}
 \label{fig:multi_view}
 \end{figure*}

\noindent \textbf{Further Qualitative Results.}
We provide further visualization results on the validation set of NuScenes (Figure \ref{fig:single_view} and Figure \ref{fig:multi_view} ) and SemanticKITTI (Figure \ref{fig:kitti}) dataset to detailedly demonstrate the panoptic segmentation ability of our network. In Figure \ref{fig:single_view}, we display the objects whose perspective projections are within the single image and compare ground-truth labels and predictions in 3D and perspective view. Our network effectively recognizes small objects (such as \textit{bicycle} and \textit{motorcycle}) and rare objects (such as \textit{trailers} and \textit{construction vehicle}). Especially in the right-construction vehicle sub-figure, our segmentation quality is slightly better than ground-truth labels at the position of robotic arms. Regarding Figure \ref{fig:multi_view}, we compare the visualization results of objects across multiple images. We primarily select challenging scenarios such as crowding (pedestrian and car) and severe occlusion (truck). For example, the truck segmentation is largely occluded by walls, which severely damages the geometric structure in LiDAR scenes and feature completeness in images. Under such conditions, our network can correctly segment most of the truck instances while missing one truck only (which is occluded by the orange construction vehicles). Moreover, for pedestrian segmentation, our network additionally segments two more occluded figures in the middle image column, although it wrongly recognizes two tiny figures as one person in the leftmost image column.

\end{subappendices}

\end{document}